%% file: main.tex
\documentclass[10pt,twocolumn,letterpaper]{article}

\usepackage[pagenumbers]{cvpr}

\usepackage[table,dvipsnames]{xcolor}
\usepackage{graphicx}
\usepackage{amsmath}
\usepackage{amssymb}
\usepackage{multirow}
\usepackage{booktabs}
\usepackage{array}
\usepackage{subcaption}
\usepackage{colortbl} 
\usepackage{graphics}

\definecolor{cvprblue}{rgb}{0.21,0.49,0.74}
\usepackage[pagebackref,breaklinks,colorlinks,citecolor=cvprblue]{hyperref}
\graphicspath{{figures/} }

\input{_macros}

\title{\methodtitle}
\author{
Wojciech Zielonka$^{1,2,3^{*}}$, 
Timur Bagautdinov$^{3}$, 
Shunsuke Saito$^{3}$,
Michael Zollhöfer$^{3}$, \\
Justus Thies$^{1,2}$,
Javier Romero$^{3}$ \\ \\
$^1$Max Planck Institute for Intelligent Systems, Tübingen, Germany
\vspace{0.1cm}\\
$^2$Technical University of Darmstadt \hspace{0.4cm} $^3$Codec Avatars Lab, Meta\\
{ \url{https://zielon.github.io/d3ga/}}
}

\begin{document}

\twocolumn[{%
\renewcommand\twocolumn[1][]{#1}%
\maketitle
\begin{center}
    \centering
    \vspace{-0.3cm}
    \captionsetup{type=figure}
    \includegraphics[width=\textwidth]{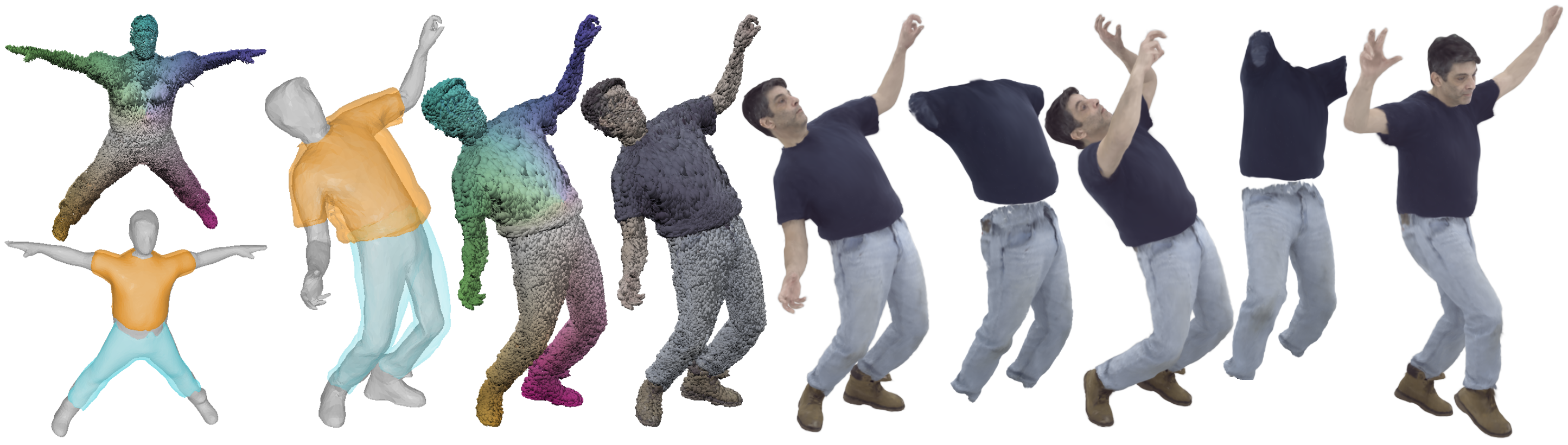}
    \vspace{-0.5cm}
    \caption{Given a multi-view video input, \model is trained to create light, drivable, photorealistic 3D human avatars. These avatars are constructed as a composition of 3D Gaussians encapsulated within tetrahedral cages. The Gaussians undergo transformation and stretching influenced by these cages, are colored using an MLP, and are rasterized into splats. By representing the drivable human as a collection of 3D Gaussian layers, we gain the ability to decompose and manipulate the avatar as needed. }
    \label{fig:teaser}
\end{center}%
}]

\let\thefootnote\relax\footnotetext{$^*$Work done while Wojciech Zielonka was an intern at Codec Avatars Lab, Meta, Pittsburgh, PA, USA}\par

\begin{abstract}
\input{0_abstract}
\end{abstract}
\vspace{-0.25cm}

\section{Introduction}
\label{sec:intro}
\input{1_introduction}

\section{Related Work}
\label{sec:related}
\input{2_related_work}

\section{Method}
\label{sec:main}
\input{3_method}

\section{Dataset}
\label{sec:data}
\input{4_data}

\section{Results}
\label{sec:results}
\input{5_results}

\section{Discussion}
\label{sec:discussion}
\input{6_discussion}

\section{Conclusion}
\label{sec:conclusion}
\input{7_conclusion} 

\input{9_acknowledgement}

{
    \small
    \bibliographystyle{ieeenat_fullname}
    \bibliography{main}
}

\clearpage
\appendix 
\input{8_appendix}

\end{document}

%% file: _macros.tex
\usepackage{comment}

\iftrue
	\definecolor{jtcolor}{RGB}{0, 0, 255}
	\newcommand\JT[1] {\emph{\textcolor{jtcolor}{JT: #1}}}
	\definecolor{wzcolor}{RGB}{0, 200, 0}
	\newcommand\WZ[1] {\emph{\textcolor{wzcolor}{WZ: #1}}}
        \definecolor{MZcolor}{RGB}{200, 0, 0}
	\newcommand\MZ[1] {\emph{\textcolor{MZcolor}{MZ: #1}}}
	\definecolor{todocolor}{RGB}{255, 125, 125}
	\newcommand\TODO[1] {\PackageWarning{}{Unprocessed todo}\emph{\textcolor{todocolor}{\textbf{TODO: #1}}}}

\else 
	\newcommand\JT[1] {}
	\newcommand\TODO[1] {}
	\newcommand\WZ[1] {}
    \newcommand\MZ[1] {}
	\newcommand\TB[1] {}
\fi 

\usepackage{float}
\usepackage{bbm}
\usepackage{import}
\usepackage{tikz}
\usepackage{lipsum}
\usepackage{bm}

\usepackage{wrapfig}

\usepackage{tabularx}
\usepackage[table]{xcolor}
\usepackage{pifont} 
\usepackage{booktabs}
\usepackage{multirow}
\usepackage{adjustbox}
\usepackage{fontawesome} 
\usepackage{amsmath} 
\usepackage{url}
\usepackage{siunitx}
\usepackage{array}
\newcolumntype{P}[1]{>{\centering\arraybackslash}p{#1}}
\newcolumntype{M}[1]{>{\centering\arraybackslash}m{#1}}

\definecolor{gold}{rgb}{1.0, 0.87, 0.0}
\definecolor{silver}{rgb}{0.75, 0.75, 0.75}
\definecolor{bronze}{rgb}{0.8, 0.5, 0.2}
\definecolor{gray}{rgb}{0.5, 0.5, 0.5}
\definecolor{lightgray}{rgb}{0.83, 0.83, 0.83}
\definecolor{lbsred}{HTML}{C87373}
\definecolor{ffdblue}{HTML}{9191FA}
\definecolor{cmarkcolor}{rgb}{0.49,0.74,0.49}
\definecolor{xmarkcolor}{rgb}{0.86,0.34,0.34}

\newcommand{\xmark}{\textcolor{xmarkcolor}{\ding{55}}}

\def\link#1{
    \ifx&#1&
        \xmark{}
    \else
        {\href{#1}{\faExternalLink}}
    \fi
}

\renewcommand{\paragraph}[1]{\vspace{1em}\noindent\textbf{#1}.}

\newcommand{\methodtitle}{Drivable 3D Gaussian Avatars\xspace}
\newcommand{\model}{D3GA\xspace}

\renewcommand{\paragraph}[1]{\smallskip\noindent\textbf{#1}}

%% file: 0_abstract.tex
\vspace{-0.25cm}
We present \methodtitle (\model), a multi-layered 3D controllable model for human bodies that utilizes 3D Gaussian primitives embedded into tetrahedral cages.
The advantage of using cages compared to commonly employed linear blend skinning (LBS) is that primitives like 3D Gaussians are naturally re-oriented and their kernels are stretched via the deformation gradients of the encapsulating tetrahedron. Additional offsets are modeled for the tetrahedron vertices, effectively decoupling the low-dimensional driving poses from the extensive set of primitives to be rendered. This separation is achieved through the localized influence of each tetrahedron on 3D Gaussians, resulting in improved optimization.
Using the cage-based deformation model, we introduce a compositional pipeline that decomposes an avatar into layers, such as garments, hands, or faces, improving the modeling of phenomena like garment sliding.
These parts can be conditioned on different driving signals, such as keypoints for facial expressions or joint-angle vectors for garments and the body.
Our experiments on two multi-view datasets with varied body shapes, clothes, and motions show higher-quality results. They surpass PSNR and SSIM metrics of other SOTA methods using the same data while offering greater flexibility and compactness.

%% file: 1_introduction.tex
Developing drivable, photorealistic human avatars is crucial for better long-distance telecommunication that provides an immersive experience to the users. 
The motion and deformations across various segments of a complex avatar's body are influenced by distinct signals, such as facial expressions and body movements. This complexity poses challenges for accurate modeling using a single layer.
Multi-layered avatars become essential to represent these different regions, ensuring proper motion and visual fidelity. Similarly, garments present challenges such as sliding, necessitating separate modeling of each clothing piece. 

Mixture of Volumetric Primitives (MVP) \cite{Lombardi2021MixtureOV} started a successful line of hybrid implementations, where volumetric primitives are embedded on the surface of the tracked mesh. This representation, despite excellent results, struggles when the provided mesh is not precise or lacks details, ultimately producing artifacts and misaligning the primitives. Similar CNN-based architectures \cite{Bagautdinov2021DrivingsignalAF, Remelli2022DrivableVA, Li2023AnimatableGL, Lombardi2021MixtureOV, Liu2021NeuralA}, do not allow for easy garment decomposition and assume a fixed amount of 3D primitives since the CNN size has to be set for the training. Furthermore, numerous methods \cite{Li2022TAVATA, Su2023NPCNP, Lombardi2021MixtureOV, Bagautdinov2021DrivingsignalAF} lack the capability of layered conditioning specific to different body parts. For example, they may not support using keypoints for the face or motion vectors for clothing like t-shirts. This is an important aspect of a holistic system that, ultimately, needs to capture speech, face, gestures, and garment motion. 
State-of-the-art drivable avatars~\cite{Remelli2022DrivableVA, Xiang2023DrivableAC} require dense input signals like RGB-D images or even multi-view camera setups at test time, which might not be suitable for low-bandwidth connections in telepresence applications.
Finally, drivable NeRFs and 3DGS avatars typically rely on LBS to transform samples between canonical and observation spaces.
However, LBS is limited by the low degree of freedom of the model, whereas cages can handle more complex non-linear motion and offer additional physical properties (e.g., stretching).

We designed our method to use a minimal set of inputs and still be competitive with the ones that require more information to train an avatar.
\model models digital humans using volumetric primitives represented as 3D Gaussians embedded into a tetrahedral cage which is naturally described by phenomenons like stretching, rotation, and scaling.
Accordingly, instead of LBS, our method builds on a classic deformation model for transforming volumes~\cite{Nieto2012Deformation}.
Specifically, by recasting cages from the canonical space into a deformed one, the 3D Gaussian covariance matrices undergo the encapsulating tetrahedral deformation transformation.
Recent advancements in incorporating physics into Gaussians \cite{xie2023physgaussian, Feng2024GaussianSD} show further promise in the context of cage usage for garment modeling by capitalizing on \cite{Chen2023ShortestPT, Macklin2021ACF}. 
Also, cages decouple the representation resolution (related to the amount of Gaussians) from the degrees of freedom present in the model ultimately allowing an effective regularization of the deformations in contrast to LBS which depends on the global bone transformations only.
In addition, we employ a compositional structure based on separate body, face, and garment cages, allowing us to model those parts independently, including localized conditioning based on different driving signals (e.g., keypoints).
We train person-specific models on nine high-quality multi-view sequences with a wide range of body shapes, motion, and clothing (not limited to tight-fitting), which later can be driven with new poses from any subject. 

\medskip\noindent
In summary, we present \methodtitle (\model) with the following contributions:
\vspace{-0.15cm}
\begin{itemize}
    \itemsep0em 
    \item A light, flexible, and composable model based on 3D Gaussian primitives driven by tetrahedral cage-based deformations which improve their body modeling properties.
    \item Localized motion conditioning which enables for instance facial expressions.
\end{itemize}

%% file: 2_related_work.tex
\model reconstructs controllable digital full-body avatars using multi-view video and joint angle motion by combining 3D Gaussian Splatting (3DGS) \cite{Kerbl20233DGS} with cage-based deformations \cite{Ju2005MeanVC, Jacobson2011BoundedBW, Huang2006SubspaceGD}. Current methods for controllable avatars rely on dynamic Neural Radiance Fields (NeRF) \cite{Mihajlovic2022KeypointNeRFGI, Park2021HyperNeRF, Park2020NerfiesDN}, point-based \cite{Zheng2022PointAvatarDP, Xu2022PointNeRFPN, Ma2021ThePO}, or hybrid representations \cite{Feng2022CapturingAA, Lombardi2021MixtureOV, Bagautdinov2021DrivingsignalAF, Zielonka2022InstantVH}, which are either slow to render or fail to correctly disentangle garments from the body, leading to poor generalization to new poses.
Recently, incorporating 3DGS into dynamic scenarios has opened new research avenues \cite{Xiang2023FlashAvatarHD, qian2023gaussianavatars, Xu2023GaussianHA, Zheng2023GPSGaussianGP, Li2023AnimatableGL}. For a thorough overview, we refer readers to state-of-the-art reports on digital avatars and neural rendering \cite{Zollhfer2018StateOT, Tewari2020StateOT, Tewari2021AdvancesIN}.


\paragraph{Dynamic Neural Radiance Fields}
NeRF \cite{Mildenhall2020NeRF} is a popular appearance model for human avatars, representing scenes volumetrically with density and color information using an MLP. Images are rendered via ray casting and volumetric integration of sample points~\cite{Kajiya1986TheRE}. Many methods have successfully applied NeRF to dynamic scenes \cite{Park2021HyperNeRF, Li2020NeuralSF, Park2020NerfiesDN, Gafni2020DynamicNR, Zielonka2022InstantVH, Prinzler2022DINERDI, Xu2022PointNeRFPN, Weng2022CVPR}, achieving high-quality results.
However, most methods treat avatars as a single layer \cite{Su2021ANeRFAN, Su2023NPCNP, Su2022DANBODA, Li2022TAVATA, Peng2020NeuralBI, Zheng2022StructuredLR, Mihajlovic2022KeypointNeRFGI}, which complicates modeling phenomena like sliding or loose garments. Methods like \cite{Feng2022CapturingAA, Feng2023DELTA} address this using a hybrid representation, combining explicit geometry from SMPL\cite{Loper2015SMPLAS} with implicit dynamic NeRF. Despite impressive garment reconstruction, these methods struggle with novel pose prediction.
TECA~\cite{Zhang2023TextGuidedGA} extends SCARF to a generative framework, enabling prompt-based generation of NeRF-based accessories and hairstyles.

\paragraph{Point-based Rendering}
Before 3DGS, many methods used point-based rendering \cite{Ma2021ThePO, Zheng2022PointAvatarDP, Su2023NPCNP} or sphere splatting~\cite{Lassner2021PulsarES}, with optimizable positions and sizes. NPC by Su et al. \cite{Su2023NPCNP} defines a point-based body model for avatar representation, but requires lengthy nearest neighbor searches during training (12 hours vs. 30 minutes for our model), making it impractical for dense multi-view datasets.
Ma et al. \cite{Ma2021ThePO} represent garments as a pose-dependent function mapping SMPL points \cite{Loper2015SMPLAS} to the clothing space. This is improved in~\cite{Prokudin2023dynamic} with a neural deformation field, but both models only address geometry, not appearance.
Zheng et al. \cite{Zheng2022PointAvatarDP} represent the upper part of an avatar as a point cloud, grown during optimization and rasterized using a differentiable renderer \cite{Wang2019DifferentiableSS}. While achieving photorealistic local results, the avatars suffer from artifacts like holes.


\paragraph{Cage-based Deformations}
Cages\cite{Nieto2012Deformation} are commonly used for geometry modeling and animation, serving as sparse proxies to control all interior points, enabling efficient deformation by manipulating only cage nodes. Yifan et al.~\cite{Wang2020NeuralCages} introduced neural cages for detail-preserving shape deformation, where a neural network rigs the source object into the target via a proxy.
Garbin et al.~\cite{Garbin2022Voltemorph} extended dynamic NeRF with tetrahedron cages to unposed ray samples based on tetrahedron intersections. This method is real-time, high-quality, and controllable, but limited to objects with local deformations like heads, and not suitable for highly articulate objects like full-body avatars.
Peng et al. used a cage to deform a radiance field in CageNeRF \cite{Peng2022CageNeRF}. While their low-resolution cages can be applied to full-body avatars, they fail to model detailed features like faces or complex deformations.


\paragraph{Time-conditioned Methods}
Playback methods \cite{wu20234dgaussians, Cao2023HexPlane, TiNeuVox, Isik2023TOG, yang2023gs4d, li2022neural3dvideosynthesis} represent a scene as a time-conditioned function that cannot be arbitrarily controlled, allowing only for a novel viewpoint synthesis while traversing the time axis. Yang et al. \cite{yang2023gs4d} extended the representation of 3DGS \cite{Kerbl20233DGS} into 4DGS, effectively incorporating time into the primitive representation. Wu et al. \cite{wu20234dgaussians} combine Gaussians with 4D neural voxels, inspired by HexPlane \cite{Cao2023HexPlane}, which achieves real-time rendering and novel-view synthesis. However, these solutions fall into a different class of algorithms compared to pose-conditioned drivable avatars, which is our goal.


\paragraph{Dynamic Gaussian Splatting}
\model is based on 3D Gaussian Splatting (3DGS) \cite{Kerbl20233DGS}, a recent alternative to NeRF for modeling neural scenes. Due to its real-time capabilities and high-quality results, 3DGS has inspired numerous follow-up papers \cite{xie2023physgaussian, Feng2024GaussianSD, Jiang2024VRGSAP, Luo2024GaussianHairHM, Xiang2023FlashAvatarHD, qian2023gaussianavatars, Xu2023GaussianHA, Zheng2023GPSGaussianGP, Zielonka2024GEM, zielonka2025synshot} in areas such as physics simulation, hair modeling, head avatars, and fluid dynamics.
Several works \cite{Li2023AnimatableGL, Pang2023ASHAG, Saito2023RelightableGC} recently introduced convolutional networks to regress Gaussian maps. Despite achieving high-quality results, fixed convolutional architectures do not allow for local conditioning or adjusting the number of Gaussians during training. These methods also use up to 23 times more parameters, causing the model size to reach almost 1 GiB. In contrast, our pipeline remains lightweight and flexible, offering garment decomposition and localized conditioning. Finally, using CNNs can slow down the pipeline to around 10 FPS \cite{Li2023AnimatableGL}, whereas our method remains real-time.

%% file: 3_method.tex
\begin{figure*}[ht!]
    \centering
    \vspace{-0.3cm}
    \includegraphics[width=\textwidth]{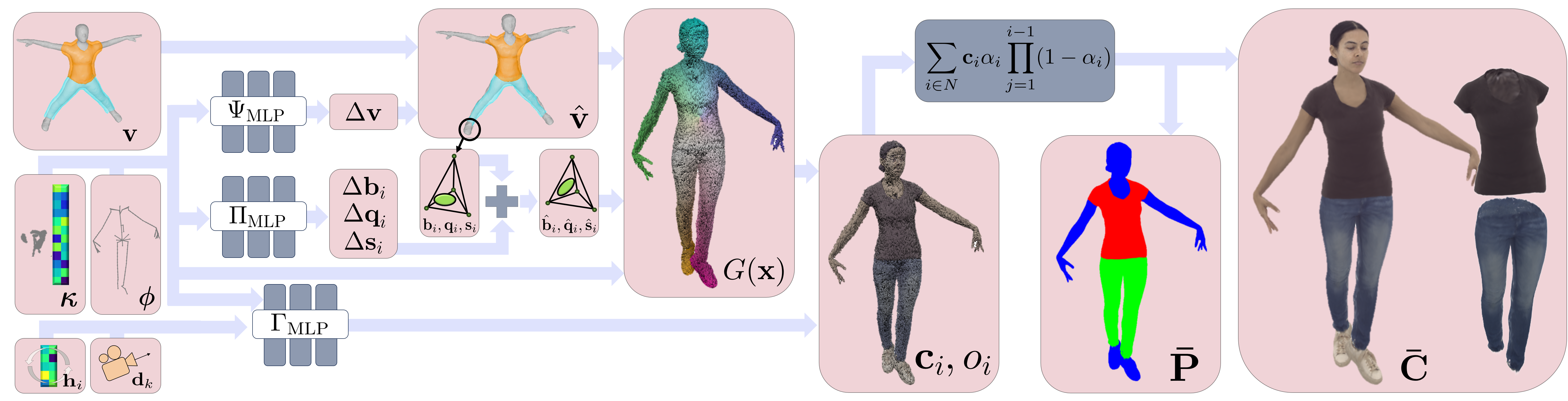}
    \vspace{-0.3cm}
    \caption{\textbf{Overview}. \model uses 3D pose $\pmb{\phi}$, face embedding $\pmb{\kappa}$, viewpoint $\mathbf{d}_k$ and canonical cage $\mathbf{v}$ (as well as auto-decoded color features $\mathbf{h}_i$) to generate the final render $\mathbf{\bar{C}}$ and auxiliary segmentation render $\mathbf{\bar{P}}$.
    The inputs in the left are processed through three networks ($\mathbf{\Psi}_\mathrm{MLP}$, $\mathbf{\Pi}_\mathrm{MLP}$, $\mathbf{\Gamma}_\mathrm{MLP}$) per avatar part to generate cage displacements $\mathbf{\Delta v}$, Gaussians deformations $\mathbf{b}_i$, $\mathbf{q}_i$, $\mathbf{s}_i$ and color/oppacity $\mathbf{c}_i$, $o_i$ respectively.
    After cage deformations transform canonical Gaussians, they are rasterized into the final images according to Eq.~\ref{formula: splatting&volume rendering}.}
    \vspace{-0.25cm}
    \label{fig:pipeline}
\end{figure*}

%
\model is built on 3DGS extended by a neural representation and tetrahedral cages to model the color and geometry of each dynamic part of the avatar, respectively.
In the following, we introduce the formulation of 3D Gaussian Splatting and give a detailed description of our method.

\subsection{3D Gaussian Splatting}
3D Gaussian Splatting (3DGS) \cite{Kerbl20233DGS} is designed for real-time novel view synthesis in multi-view static scenes. 
Their rendering primitives are scaled 3D Gaussians \cite{Wang2019DifferentiableSS, Kopanas2021PointBasedNR} with a 3D covariance matrix $\mathbf{\Sigma}$ and mean $\mathbf{\mu}$:
\begin{equation}
\label{formula: gaussian's formula}
    G(\mathbf{x})=e^{-\frac{1}{2}(\mathbf{x-\mu})^T\mathbf{\Sigma}^{-1}(\mathbf{x-\mu})}.
\end{equation}
To splat the Gaussians, Zwicker et al. \cite{Zwicker2001SurfaceS} define the projection of 3D Gaussians onto the image plane as:
\begin{equation}
    \label{formula: covariance projection}
    \mathbf{\Sigma}^{\prime} = \mathbf{A}\mathbf{W}\mathbf{\Sigma} \mathbf{W}^T\mathbf{A}^T,
\end{equation}
where $\mathbf{\Sigma}^{\prime}$ is a covariance matrix in 2D space, $\mathbf{W}$ is the view transformation, and $\mathbf{A}$ is the Jacobian of the affine approximation of the projective transformation.
During optimization, enforcing the positive semi-definiteness of the covariance matrix $\mathbf{\Sigma}$ is challenging. To avoid this, Kerbl et al. \cite{Kerbl20233DGS} use an equivalent formulation of a 3D Gaussian as a 3D ellipsoid parameterized with a scale $\mathbf{S}$ and rotation $\mathbf{R}$:
\begin{equation}
\label{formula: covariance decomposition}
    \mathbf{\Sigma} = \mathbf{R}\mathbf{S}\mathbf{S}^T\mathbf{R}^T.
\end{equation}
3DGS uses spherical harmonics \cite{Ramamoorthi2001AnER} to model the view-dependent color of each Gaussian. In practice, appearance is modeled with an optimizable 48 elements vector representing four bands of spherical harmonics.
\subsection{Body Cage Creation}

\begin{wrapfigure}{l}{0.55\columnwidth}
\centering
\setlength{\unitlength}{0.1\columnwidth}
\begin{picture}(3, 3.6)
\put(-1.1, -0.3){\includegraphics[width=0.25\textwidth]{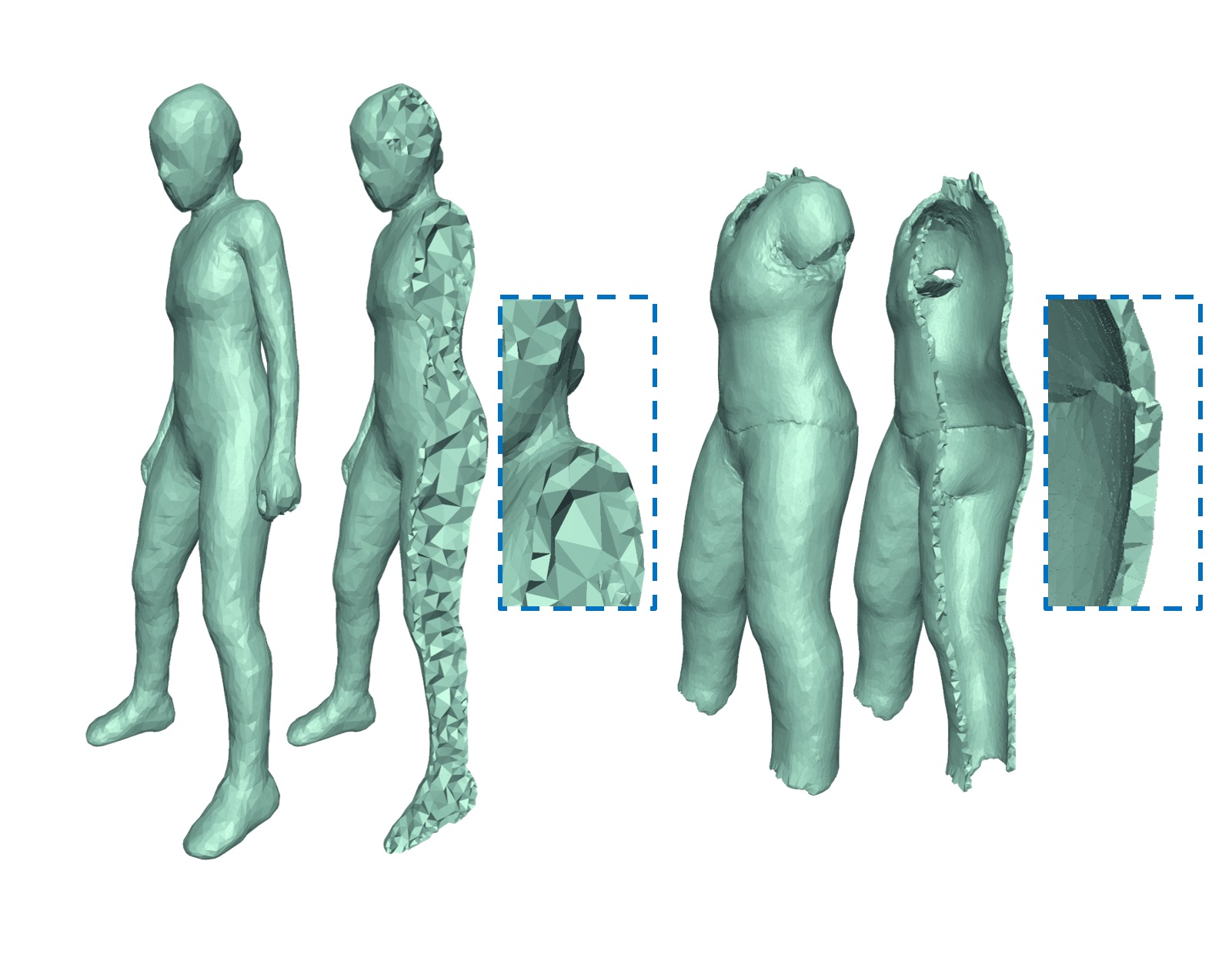}}
\put(-0.1, 3.7){\small{Body}}
\put(1.8, 3.7){\small{Garment}}
\end{picture}
\vspace{-0.35cm}
\caption{\model uses a tetrahedral mesh for deformation modeling.}
\vspace{-0.35cm}
\label{fig:tet_mesh}
\end{wrapfigure}

To deform 3D Gaussians, we utilize tetrahedron cage-based deformations as a coarse proxy for the body, face, and individual garments. Unlike a triangle, which is two-dimensional, a tetrahedron is a polyhedron with four triangular faces (A, B, C, D), providing a three-dimensional structure. The volume of a tetrahedron can be calculated using the scalar triple product of vectors, which enables precise control and deformation of the enclosed 3D Gaussians. The volume $V$ is given by:
\begin{equation}
V = \frac{1}{6} \left| \textbf{AB} \cdot (\textbf{AC} \times \textbf{AD}) \right| 
\end{equation}
where $\textbf{AB, AC, AD}$ are edges of tetraherdon. This property allows us to compute the deformation gradient similarly to Sumner et al. \cite{Sumner2004DeformationTF} and transfer it to the Gaussian covariance matrix (Equation \ref{formula: deformation gradient closed form}), see Supp. Mat, for more details.

To create a cage per garment, we segment all images of a single time instance using an EfficientNet~\cite{Tan2019EfficientNet} backbone with PointRend~\cite{Kirillov2020PointRend} refinement, trained on a corpus of similar multi-view captures.
The per-image 2D segmentation masks are projected onto a body mesh $\hat{\mathbf{M}}$ to obtain per-triangle labels (body, upper, lower). 
To get the mesh $\hat{\mathbf{M}}$, we fit a low-resolution LBS model to a single 3D scan of the subject and then fit such model to the segmented frame by minimizing the distance to the 3D keypoints, extracted with an EfficientNet trained on similar captures.
We transform the body mesh into canonical space with LBS and divide it into body part templates $\mathbf{M}_k$.
The garment meshes are additionally inflated by 1-3 cm along the vertex normals. Afterward, we run a voxelization of the meshes and subsequently extract the mesh using the marching cubes algorithm \cite{Lorensen1987MarchingCubes}.
After that, we use TetGen \cite{Si2013TetGenAQ} to turn the unposed meshes $\mathbf{M}_k$ into tetrahedral meshes $\mathbf{T}_k$.
Consequently, cages for garments are hollow, containing only their outer layer, while the body cage is solid (Figure \ref{fig:tet_mesh}). The face cage is composed of the body tetrahedra which contains triangles defined as the face on the LBS template.
The cage nodes are deformed according to LBS weights transferred from the closest vertex in $\mathbf{M}_k$.

\subsection{Cage Deformation Transfer}

While classic cage methods typically deform the volume according to complex weight definitions ~\cite{Ju2005MeanVC, Jacobson2011BoundedBW, Joshi2007TOG}, using linear weights works well in practice when cage cells are small, making it easier to integrate into an end-to-end training system.
Specifically, we define $\mathbf{v}_{ij}$ as the vertices of tetrahedron $i$ in canonical space, any point $\mathbf{x}$ inside this tetrahedron can be defined by its barycentric coordinates $b_j$:
\begin{equation}
\label{formula: barys}
\mathbf{x}=\sum_{j=1}^4 b_j \mathbf{v}_{ij}.
\end{equation}
\noindent
Each Gaussian 3D mean $\bm{\mu} = \mathbf{x}$ is obtained as a linear combination of learnable barycentric coordinates $b_j$ and tetrahedron vertices $\mathbf{v}_{ij}$. 
When the tetrahedra are transformed to posed space according to $\mathbf{\hat{v}}_{ij} = \mathrm{LBS}(\mathbf{v}_{ij}, \pmb{\phi}, \mathbf{w}_{ij})$, where $\pmb{\phi}$ is the pose and $\mathbf{w}_{ij}$ are the blendweights, the same linear relation holds $\hat{\mathbf{x}}=\sum_{j=1}^4 b_j \hat{\mathbf{v}}_{ij}$. To leverage the cage volume properties (rotation, sheer, and scaling), we use the deformation gradient \cite{Sumner2004DeformationTF}:
\begin{align}
\label{formula: deformation gradient closed form}
\mathbf{J}_i\mathbf{E}_i &= \hat{\mathbf{E}}_i,\\
\label{formula: deformation gradient closed form}
\mathbf{J}_i &= \hat{\mathbf{E}}_i \mathbf{E}_i^{-1},
\end{align}
where $\hat{\mathbf{E}}_i \in \mathbb{R}^{3\times3}$ and $\mathbf{E}_i \in \mathbb{R}^{3\times3}$ contain three edges from tetrahedron $i$ defined in deformed and canonical spaces, respectively.
The gradient $\mathbf{J}_i$ is used to transform the kernel of each Gaussian $i$ (Eq \ref{formula: deformation gradient covariance}). See Supp. mat for more details.
\subsection{Drivable Gaussian Avatars}

We initialize a fixed number of Gaussians, whose 3D means $\bm{\mu}$ are sampled on the surface of $\hat{\mathbf{M}}$. However, we are not limited to the fixed amount of Gaussians allowing for cloning or densification if needed.
The rotation of each Gaussian is initialized so that the first two axes are aligned with the triangle surface and the third one with the normal: this is a good approximation for a smooth surface.
The scale is initialized uniformly across a heuristic range depending on inter-point distances as in~\cite{Kerbl20233DGS}.
We assign each sampled position $\mathbf{x}$ to the intersecting tetrahedron and compute its barycentric coordinates $\mathbf{b} \in \mathbb{R}^{4}$.
To deform the tetrahedron volume, we incorporate the deformation gradient $\mathbf{J}$ defined in Eq.~\ref{formula: deformation gradient closed form} into the Gaussian covariance matrix from Eq.~\ref{formula: covariance decomposition}. 

This is an important step as the deformation gradient $\mathbf{J}$ encapsulates many phenomena that we want to model, for instance, rotation, stretching, and sheering. To correctly transfer the deformation to 3D Gaussian primitives, we apply it to the covariance matrix $\mathbf{\Sigma}$, effectively modeling the 3DGS ellipsoids depending on the shape deformation from the canonical space into deformed one. Thus, the final covariance matrix passed to the rasterizer is denoted as:
\begin{equation}
\label{formula: deformation gradient covariance}
\hat{\mathbf{\Sigma}} = \mathbf{J}_i\mathbf{\Sigma}\mathbf{J}_i^{T},
\end{equation}
where $\mathbf{J}_i$ is the deformation gradient of the tetrahedron containing the 3D mean of the Gaussian with covariance $\mathbf{\Sigma}$.
This way, we transfer the deformation into the Gaussians, improving modeling phenomena like garment stretching.
Each part of the avatar (the garment, body, or face) is controlled by a separate GaussianNet $\mathbb{G}_{\text{Net}} = \{ \Gamma_{\text{MLP}}, \Pi_{\text{MLP}}, \Psi_{\text{MLP}}\}$ which is defined as a set of small specialized multi-layer perceptrons (MLP) parametrized as:
\begin{equation}
\begin{split}
\label{formula: mlps}
\Psi_{\text{MLP}}&: \{\pmb{\phi}, \text{enc}_{pos}(\mathbf{v})\} \rightarrow \Delta\mathbf{v}, \\
\Pi_{\text{MLP}}&: \{\pmb{\phi}, \mathbf{b}_i, \mathbf{q}_i, \mathbf{s}_i\} \rightarrow \{\Delta\mathbf{b}_i, \Delta\mathbf{s}_i, \Delta\mathbf{q}_i\}, \\
\Gamma_{\text{MLP}}&: \{\pmb{\phi}, \text{enc}_{view}(\mathbf{d}_k), \mathbf{h}_i, \mathbf{f}_j\} \rightarrow \{\mathbf{c}_i, o_i\}. \\
\end{split}
\end{equation}
All the networks take joint angles $\pmb{\phi}$ (or face encodings $\pmb{\kappa}$ for the face networks) as inputs, in addition to network-specific conditioning.
The cage node correction network $\Psi_{\text{MLP}}$ takes positional encodings~\cite{Mildenhall2020NeRF} for all canonical vertices to transform them into offsets of the cage node positions similar to SMPL~\cite{Loper2015SMPLAS} pose-correctives.
To adapt our representation further to the pose, the Gaussian correction network $\Pi_{\text{MLP}}$ takes the canonical Gaussian parameters (barycentric coordinates $\mathbf{b}_i \in \mathbb{R}^{4}$, rotation $\mathbf{q}_i \in \mathbb{R}^{4}$ and scale $\mathbf{s}_i \in \mathbb{R}^{3}$) to predict corrections of those same parameters.
These two networks are necessary to capture high-frequency details outside the parametric transformation.

The shading network $\Gamma_{\text{MLP}}$ transforms encoded view direction and initial color into final color and opacity, $\mathbf{c}_i, o_i$. Unlike 3DGS, we use a pose-dependent color representation to model self-shadows and wrinkles in garments. The view angle is projected onto the first four spherical harmonics bands $\text{enc}_{pos}(\cdot)$, while the initial color is an auto-decoded feature vector $\mathbf{h}_i$ \cite{Park2019Deepsdf}.
Additionally, the face region utilizes face embeddings $\pmb{\kappa}$ as input instead of pose $\pmb{\phi}$. This adaptability stems from our model's composability and holds the potential for extension to other regions, such as hair, shoes, or hands. A small auxiliary MLP regresses $\pmb{\kappa}$ based on 150 3D keypoints $\mathbf{k}$ normalized by their training mean and standard deviations. This effectively enables us to model facial expressions.

Finally, we also add an embedding vector with the time frame of the current sample \cite{MartinBrualla2020NeRFIT}.
This allows \model to explain properties that cannot be modeled (e.g., cloth dynamics) from our input, effectively avoiding excessive blur due to averaging residuals.
During testing, the average training embedding is used.
\subsection{Training Objectives}
As in 3DGS \cite{Kerbl20233DGS}, we define the color $\mathbf{\bar{C}}$ of pixel $(u,v)$:
\begin{equation}
\label{formula: splatting&volume rendering}
    \mathbf{\bar{C}}_{u,v} = \sum_{i\in \mathcal{N}}\mathbf{c_i} \alpha_i \prod_{j=1}^{i-1} (1-\alpha_i),
\end{equation}
where $\mathbf{c_i}$ is the color predicted by $\Gamma_{\text{MLP}}$, which replaces the spherical harmonics in 3DGS.
$\alpha_i$ is computed as the product of the Gaussian density in Eq.~\ref{formula: gaussian's formula} with covariance matrix $\mathbf{\Sigma}^{\prime}$ from Eq.~\ref{formula: covariance projection} and the learned per-point opacity $o_i$ predicted by $\Gamma_{\text{MLP}}$.
The sum is computed over set $\mathcal{N}$, the Gaussians with spatial support on $(u,v)$.
The primary loss in \model is a weighted sum of three different color losses applied to the estimated image $\mathbf{\bar{C}}$ and the ground truth RGB image $\mathbf{C}$:
\begin{equation}
\label{formula: color}
\mathcal{L}_{Color} = (1 - \omega) \mathcal{L}_1 + \omega\mathcal{L}_{\text{D-SSIM}} + \zeta\mathcal{L}_{\text{VGG}}, \\
\end{equation}
where $\omega=0.2$, $\zeta=0.005$ (after 400k iterations steps and zero otherwise), $\mathcal{L}_{\text{D-SSIM}}$ is a structural dissimilarity loss, and $\mathcal{L}_{\text{VGG}}$ is the perceptual VGG loss.

To encourage correct garment separation, we introduce a garment loss.
Since each Gaussian $i$ is statically assigned to a part, we define $\mathbf{p_i}$ as a constant-per-part color and consequently render $\mathbf{\bar{P}}$ by replacing $\mathbf{c_i}$ by $\mathbf{p_i}$ in Eq.~\ref{formula: splatting&volume rendering}.
Then, we compute the $\mathcal{L}_1$ norm between predicted parts $\mathbf{\bar{P}}$ and ground truth segmentations $\mathbf{P}$, $\mathcal{L}_{Garment} = \mathcal{L}_1(\mathbf{\bar{P}},\mathbf{P})$.
Moreover, we are using the Neo-Hookean loss based on Macklin et al. \cite{Macklin2021ACF} to enforce the regularization of the predicted tetrahedra for the regions with low supervision signal:
\begin{equation}
    \label{formula: neo}
    \mathcal{L}_{Neo} = \frac{1}{N}\sum_{i=0}^{N}\frac{\lambda}{2}\left(\text{det}(\mathbf{J}_i) - 1\right)^2 + \frac{\mu}{2}\left(\text{tr}(\mathbf{J}_i^T\mathbf{J}_i) - 3\right),
\end{equation}
where $\mathbf{J}_i$ denotes the deformation gradient between a canonical and a deformed tetrahedron (Eq.~\ref{formula: deformation gradient closed form}), $N$ is the total number of tetrahedrons, and $\lambda$ and $\mu$ are the Lam\'e parameters~\cite{Macklin2021ACF}.
The overall loss is defined as:
\begin{equation}
\mathcal{L} = \nu\mathcal{L}_{Color} + \nu\mathcal{L}_{Garment} + \tau\mathcal{L}_{Neo},
\end{equation}
where $\nu=10$ and $\tau=0.005$ balance the different losses.

We implemented \model based on the differentiable 3DGS renderer \cite{Kerbl20233DGS}. 
The networks $\Pi_{\text{MLP}}$,$\Psi_{\text{MLP}}, \Gamma_{\text{MLP}}$ have three hidden layers with 128 neurons and ReLU activation functions.
In our experiments, we train the networks for 700k (Ours) and 400k (ActorsHQ) steps with a multi-step scheduler with a decay rate of $0.33$, a batch size of one, and using the Adam optimizer \cite{Kingma2014AdamAM} with a learning rate set to $5e-4$.
We ran all experiments on a single Nvidia V100 GPU with $1024\times667$ images.
When ground truth poses are not available, as in the case of ActorsHQ \cite{Isik2023TOG}, we additionally refine poses regressed from keypoints during avatar training and during the test time, and optionally projected them onto PCA basis computed from the training set.

\begin{figure*}[ht!]
    \vspace{-1.5cm}
    \setlength{\unitlength}{0.1\textwidth}
    \begin{picture}(10, 7)
    \put(0.5, 0.0){\includegraphics[width=0.9\textwidth]{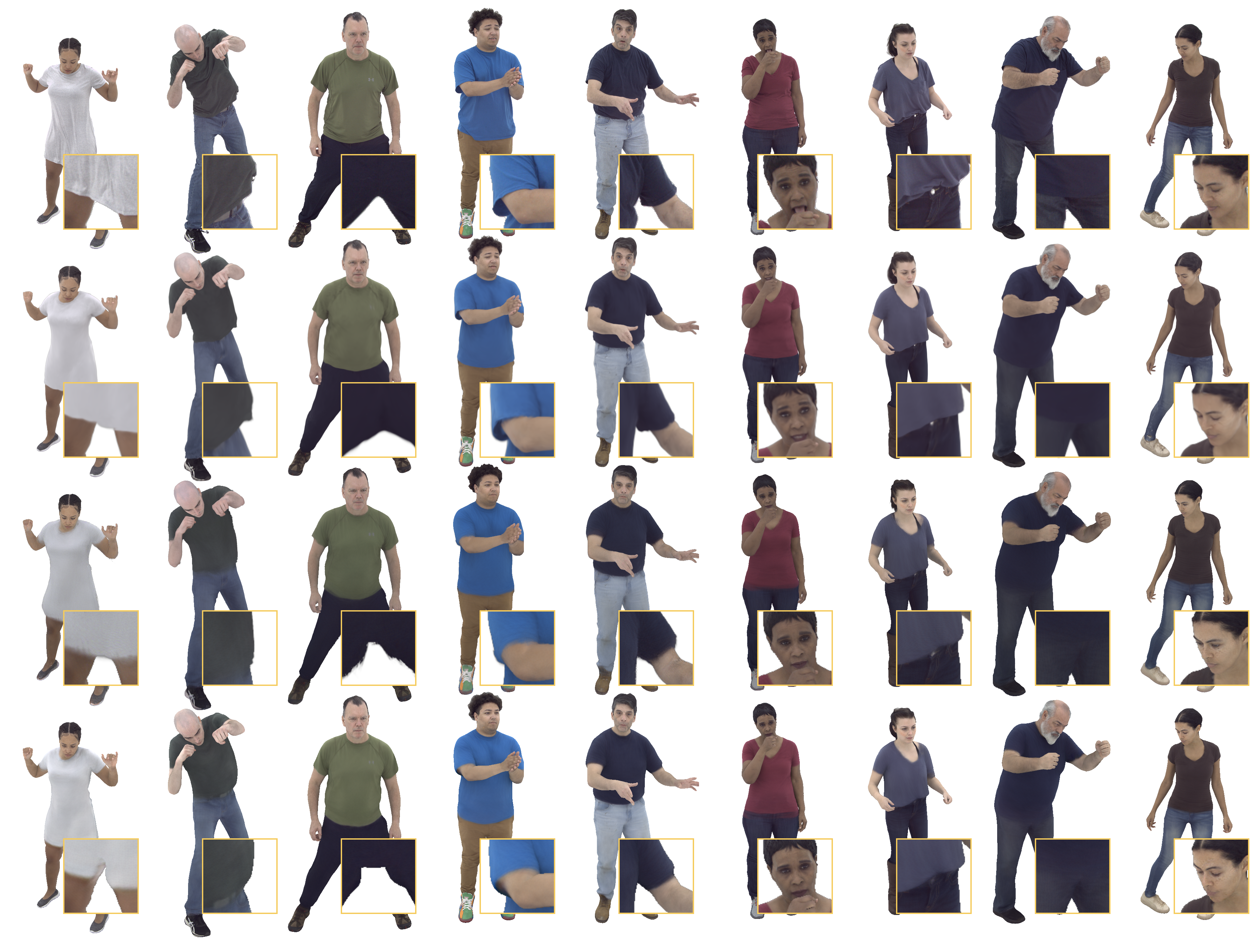}}
    \put(0.3, 5.2){\rotatebox{90}{Ground Truth}}
    \put(0.3, 4.0){\rotatebox{90}{Ours}}
    \put(0.3, 2.2){\rotatebox{90}{MVP\cite{Lombardi2021MixtureOV}}}
    \put(0.3, 0.5){\rotatebox{90}{BD \cite{Bagautdinov2021DrivingsignalAF}}}
    \end{picture}
    \vspace{-0.6cm}
    \caption{Qualitative comparisons show that \model models facial expressions and garments better than other SOTA approaches. Especially regions with loose garments like skirts or sweatpants.}
    \label{fig:main_results}
    \vspace{-0.3cm}
\end{figure*}

%% file: 4_data.tex
Our dataset comprises nine subjects performing various motions, observed by 200 cameras. We use 12,000 frames for training (at 10 FPS) and 1,500 for testing (at 30 FPS). Images were captured at a resolution of $4096 \times 2668$ in a multi-view studio with synchronized cameras and downsampled to $1024 \times 667$ to reduce computational cost. We utilize 2D segmentation masks, RGB images, keypoints, and 3D joint angles for training, as well as a single registered mesh to create our template $\hat{\mathbf{M}}$. Of the nine subjects, data for four is publicly available through the Goliath-4 dataset release \cite{martinez2024codec}.

%% file: 5_results.tex
We evaluate and benchmark our method w.r.t. five state-of-the-art multiview-based solutions \cite{Bagautdinov2021DrivingsignalAF, Lombardi2021MixtureOV, hu2024gaussianavatar, qian20233dgsavatar, Li2023AnimatableGL}. 
We compare \model to the mesh-based full-body avatar methods BodyDecoder (\textbf{BD}) \cite{Bagautdinov2021DrivingsignalAF} and MVP-based avatars \cite{Remelli2022DrivableVA, Lombardi2021MixtureOV} evaluated on our dataset. 

Additionally, we evaluated \model on the ActorsHQ dataset \cite{Isik2023TOG} using a significantly smaller number of cameras (40). We compare to SOTA pose-conditioned 3DGS avatar methods, including Animatable Gaussians (\textbf{AG}) \cite{Li2023AnimatableGL}, 3DGS-Avatar \cite{qian20233dgsavatar}, and Gaussian Avatar (\textbf{GA}) \cite{hu2024gaussianavatar} which were trained on the same multiview data.

Please note that our method, along with 3DGS-Avatar and GA, represents a lightweight class of MLP-based algorithms, utilizing up to 10 million parameters. In contrast, the CNN-based MVP, BD, and AG \cite{Li2023AnimatableGL} which in this case uses approximately $23$ times more parameters (230 million).
\subsection{Image Quality Evaluation}
Our model is evaluated using SSIM, PSNR, and the perceptual metric LPIPS \cite{Zhang2018TheUE}, with random color backgrounds. For the ActorsHQ evaluation, we utilized SMPL-X fittings obtained through OpenPose \cite{OpenPose} and scan-to-mesh optimization.
Table \ref{tab:color_error_socio} shows that our method achieves the best PSNR and SSIM on our dataset compared to MVP \cite{Lombardi2021MixtureOV} and BD \cite{Bagautdinov2021DrivingsignalAF}. Furthermore, on the ActorsHQ dataset, \model outperforms other Gaussian Avatar methods in terms of PSNR and SSIM. However, similar to previous evaluations, our method lacks sharpness due to its much smaller size compared to the CNN-based architecture of AG \cite{Li2023AnimatableGL}.
Moreover, our approach allows us to decompose avatars into drivable layers, unlike other volumetric methods. Each separate garment layer can be controlled solely by skeleton joint angles, without requiring specific garment registration modules as in \cite{Xiang2021ModelingCA}.

\input{tables/color_error_socio}

\begin{figure*}[t!]
    \vspace{-0.6cm}
    \setlength{\unitlength}{0.1\textwidth}
    \begin{picture}(10, 10)
    \put(0.5, 0.0){\includegraphics[width=0.9\textwidth]{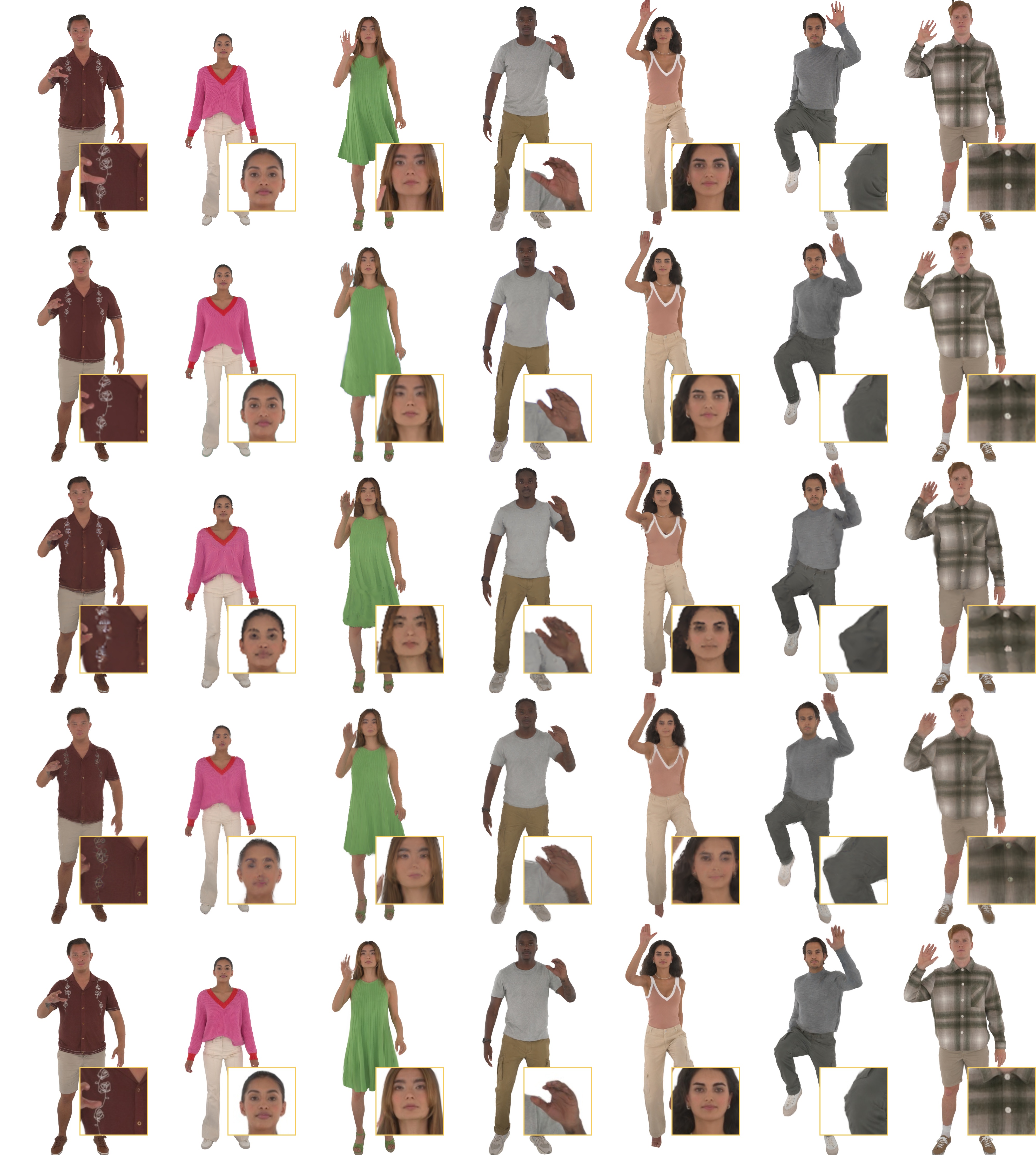}}
    \put(0.3, 8.3){\rotatebox{90}{Ground Truth}}
    \put(0.3, 6.7){\rotatebox{90}{Ours}}
    \put(0.3, 4.25){\rotatebox{90}{3DGS-Avatar \cite{qian20233dgsavatar}}}
    \put(0.3, 2.2){\rotatebox{90}{Gaussian Avatar \cite{hu2024gaussianavatar}}}
    \put(0.3, 0.1){\rotatebox{90}{Animitable GA \cite{Li2023AnimatableGL}}}
    \end{picture}
    \vspace{-0.4cm}
    \caption{ActorsHQ \cite{Isik2023TOG} comprises challenging garments that contain high-frequency patterns. Our method despite its small size can capture it and performs the best in terms of PSNR and SSIM, ranking second only in terms of sharpness to AG \cite{Li2023AnimatableGL}, which presents very sharp results due to the powerful StyleUNet \cite{wang2023styleavatar}.}
    \label{fig:main_results}
\end{figure*}

\begin{figure}[ht!]
\centering
\vspace{0.1cm}
\includegraphics[width=0.45\textwidth]{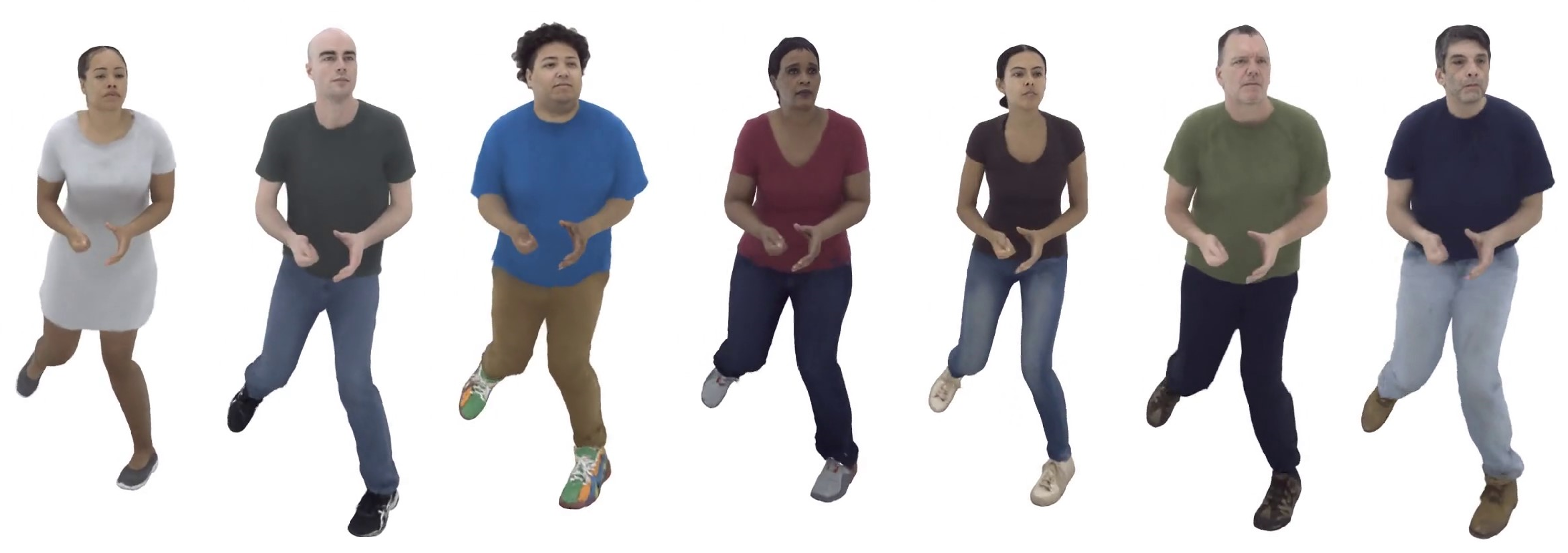}
\vspace{-0.4cm} 
\caption{\model enables motion transfer showing good generalizability while preserving each avatar's high-quality details.}
\label{fig:transfer}
\end{figure}

\subsection{Ablation Studies}

\begin{figure}[h!]
\centering
\setlength{\unitlength}{0.1\textwidth}
\begin{picture}(3, 3.3)
\put(-1.1, 0){\includegraphics[width=0.5\textwidth]{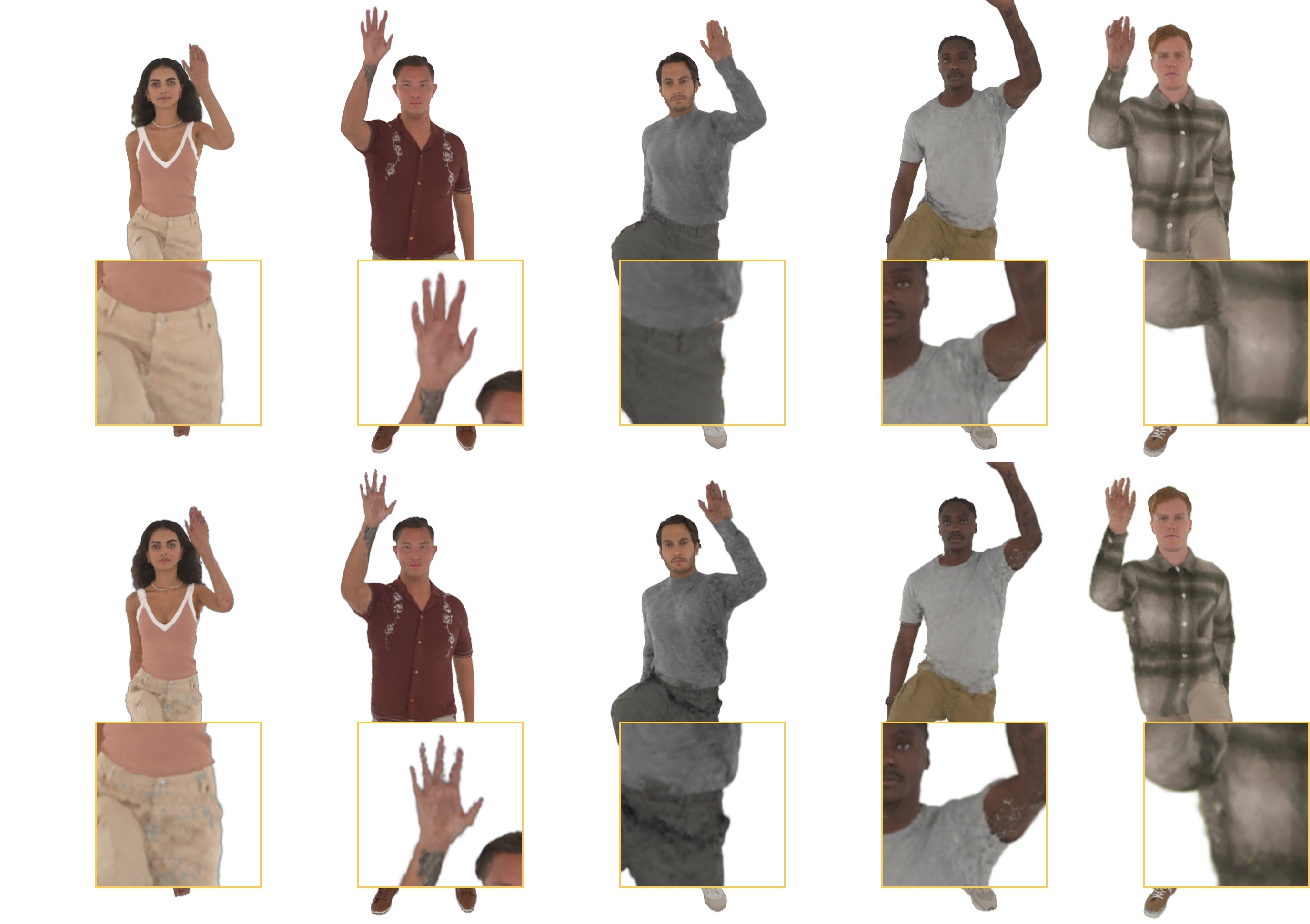}}
\put(-1.0, 2){\rotatebox{90}{Tetrahedrons}}
\put(-1.0, 0.25){\rotatebox{90}{Triangles}}
\end{picture}
\vspace{-0.2cm}
\caption{Gaussian primitives embedded in triangles, compared to tetrahedrons, produce more artifacts, resulting in small holes and reduced sharpness that is reflected in the LIPIS score, which drops from $0.0648$ to $0.0703$.}
\label{fig:cage-tris}
\vspace{-0.4cm}
\end{figure}

\paragraph{Importance of cage deformations}
We replaced tetrahedrons with triangles to emphasize the crucial role of cage deformation gradients in transforming Gaussians. We modified Eq \ref{formula: barys} such that 3D means are obtained through the barycentric coordinates of triangles $\mathbf{b} \in \mathbb{R}^{3}$ instead of tetrahedrons $\mathbf{b} \in \mathbb{R}^{4}$. The rest of the pipeline remains unchanged, with MLPs computing the same corrective terms as our cage-based model. Since triangles do not provide volume, we disabled the application of the cage deformation gradient $\mathbf{J}$, but the Gaussians are still modeled by the predicted residuals w.r.t. the canonical space. Figure \ref{fig:cage-tris} shows that the triangle-based approach does not stretch the primitives correctly, creating holes and artifacts which demonstrates the importance of using cages for deformation.

\begin{figure}[b!]
    \centering
    \vspace{-0.1cm}
    \setlength{\unitlength}{0.1\textwidth}
    \begin{picture}(3, 2)
    \put(-0.5, 0){\includegraphics[width=0.4\textwidth]{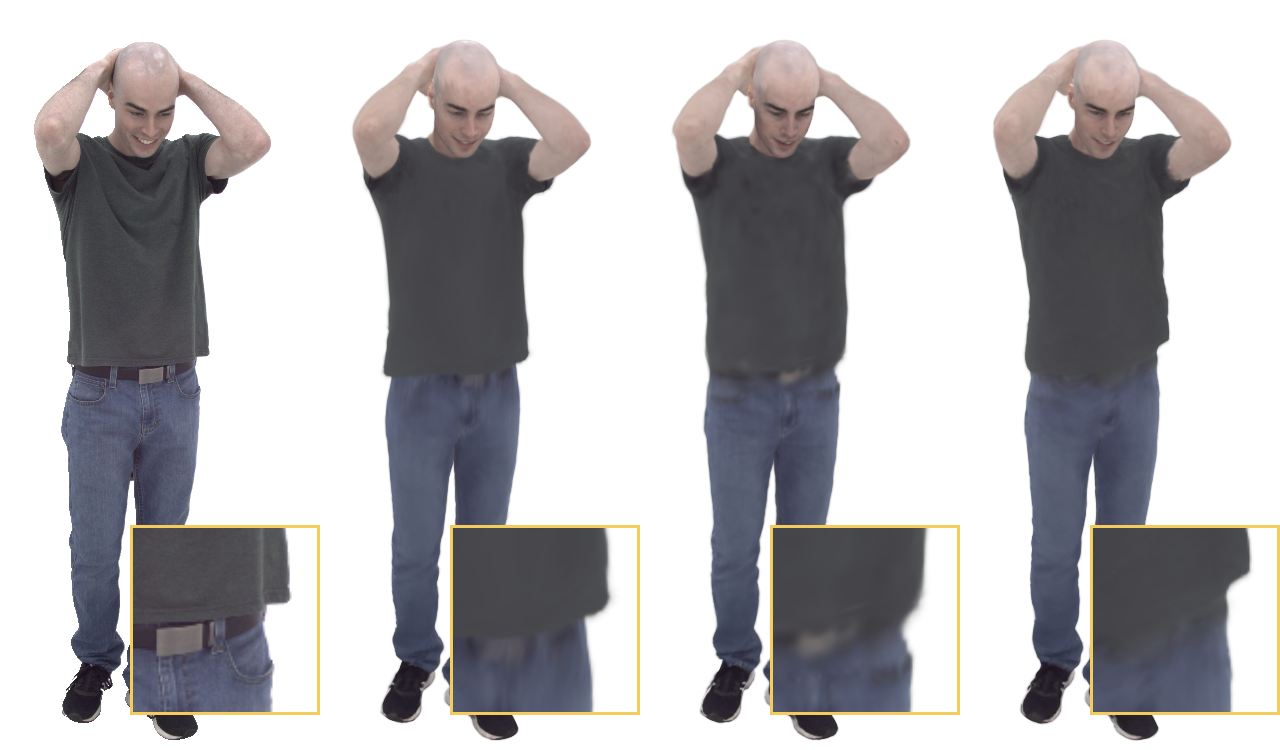}}
    \put(-0.6, -0.3){Ground Truth}
    \put(0.8, -0.3){Ours}
    \put(1.4, -0.3){w/o $\mathcal{L}_{Garment}$}
    \put(2.8, -0.3){Single Layer}
    \end{picture}
    \vspace{0.5cm}
    \caption{Ablation of \model: shape smoothness without $\mathcal{L}_{Garment}$, and sliding artifacts with a single layer representation.}
    \label{fig:color_error_ablation}
\end{figure}

\paragraph{Garment loss}
The garment loss $\mathcal{L}_{Garment}$ (Fig. \ref{fig:color_error_ablation}) serves two primary purposes: it improves garment separation and reduces erroneously translucid regions.
We can observe qualitatively that regions between garments' boundaries without the regularizer are blurry and have erroneous opacity, see supp. mat.

\paragraph{Single layer avatar}
\model supports a single-layer training for the garment and body, which struggles to model proper garment sliding. The results are presented in the last column of Fig.~\ref{fig:color_error_ablation}.
It can be observed that the edges between the T-shirt and jeans are over-smoothed.

\paragraph{Size and compactness}
Our model offers an optimal balance between quality and model size, making it both compact and easily portable. This lightweight representation sets \model apart from much larger and more cumbersome models like AG \cite{Li2023AnimatableGL}. As shown in Table \ref{tab:model_size}, \model is similar in size to other methods, yet it delivers superior quality compared to models in the same category. This makes \model an attractive choice for telepresence applications, where both efficiency and performance are crucial.

\input{tables/model_size}

%% file: tables/color_error_socio.tex
\begin{table}[h!]
    \vspace{-0.3cm}
    \centering
    \resizebox{1.0\linewidth}{!}{
    \begin{tabular}{l|l|rrr}
    Dataset & Method & PSNR $\uparrow$ & LPIPS $\downarrow$ & SSIM $\uparrow$ \\
    \midrule
    \multirow{3}{*}{Ours} 
    & Ours & \textbf{30.634} & 0.054 & \textbf{0.964} \\
    & MVP \cite{Lombardi2021MixtureOV} & 28.795 & 0.051 & 0.955 \\
    & BD \cite{Bagautdinov2021DrivingsignalAF} & 29.918 & \textbf{0.044} & 0.959 \\
    \midrule
    \multirow{4}{*}{ActorsHQ}
    & Ours & \textbf{26.562} & 0.065 & \textbf{0.944} \\
    & GA \cite{hu2024gaussianavatar} & 24.731 & 0.088 & 0.933 \\
    & 3DGS-Avatar \cite{qian20233dgsavatar} & 21.709 & 0.082 & 0.915 \\
    & AG \cite{li2024animatablegaussians} & 26.454 & \textbf{0.055} & 0.937 \\
    \bottomrule
    \end{tabular}
    }
    \caption{Our method scores the best in terms of PSNR and SSIM compared to BD \cite{Bagautdinov2021DrivingsignalAF} and MVP \cite{Lombardi2021MixtureOV} on our dataset. \model is the best among MLP-based avatars, ranking only second in terms of sharpness compared to AG, which uses a CNN-based architecture.}
    \label{tab:color_error_socio}
    \vspace{-0.5cm}
\end{table}

%% file: tables/model_size.tex
\begin{table}[t!]
    \centering
    \resizebox{0.8\linewidth}{!}{
    \begin{tabular}{l|rr}
    Method & \#parameters (M) & size (MiB) \\
    \midrule
    Ours & 9 & 45 \\
    GaussianAvatar \cite{hu2024gaussianavatar} & 7 & 59 \\
    3DGS-Avatar \cite{qian20233dgsavatar} & 6 & 57 \\
    AG \cite{li2024animatablegaussians} & 232 & 862 \\
    \bottomrule
    \end{tabular}
    }
     \caption{Model compactness. \model offers the best tradeoff between quality and model size.}
    \label{tab:model_size}
    \vspace{-0.5cm}
\end{table}

%% file: 6_discussion.tex
While \model shows better quality and competitive rendering speed w.r.t. the state of the art, there are still particular challenges.
High-frequency patterns, like stripes, may result in blurry regions.
One way of improving image quality would be using a variational autoencoder to regress Gaussian parameters per texel of a guide mesh similar to \cite{Lombardi2021MixtureOV, Li2023AnimatableGL}.
Despite using the $\mathcal{L}_{Garment}$ loss, self-collisions for loose garments are still challenging, and the sparse controlling signal does not contain enough information about complex wrinkles or self-shadowing. A potential solution to solve self-penetration would be to incorporate explicit collision detection \cite{Chen2023ShortestPT} for the tetrahedrons.
An exciting follow-up work direction would be replacing the appearance model in \model with a relightable one.
D3GA is currently limited to model photorealistic avatars for a few consenting subjects captured in a dense multi-view capture device. While this limits the potential misuse of the technology of driving somebody else's avatar without their consent, it needs to be addressed in future work.
%
%
In conclusion, it's worth noting that the \model offers significant flexibility and can be customized for particular applications. For instance, one could employ additional Gaussians to capture high-frequency detail or opt to eliminate garment supervision, particularly if precise cage geometry decomposition isn't necessary.

%% file: 7_conclusion.tex
We have proposed \model, a novel approach for reconstructing multi-layered animatable human avatars using tetrahedral cages embedded with 3D Gaussians. 
To transform the rendering primitives from canonical to deformed space, we directly apply the deformation gradient to the 3D Gaussian parametrization, enabling improved avatar modeling. 
Our method's compositional approach enables various forms of localized conditioning, such as using keypoints for facial expressions, and can be extended to other regions like hair, hands, or shoes. 
This capability is essential for creating holistic avatars driven by diverse input signals. 
We have demonstrated high-quality results that surpass state-of-the-art methods with similar model architectures, all while maintaining a lightweight, real-time, and compact approach.

%% file: 9_acknowledgement.tex
\paragraph{Acknowledgement}
The authors thank the International Max Planck Research School for Intelligent Systems (IMPRS-IS) for supporting WZ. We also want to thank Giljoo Nam for the help with Gaussian visualizations, and Anka Chen for very useful conversations about tetrahedrons.

%% file: 8_appendix.tex
\twocolumn[{%
\renewcommand\twocolumn[1][]{#1}%
\begin{center}
    \textbf{\Large{\methodtitle \\ -- Supplemental Document --}}
    \centering
    \vspace{0.5cm}
    \captionsetup{type=figure}
    \includegraphics[width=0.85\textwidth]{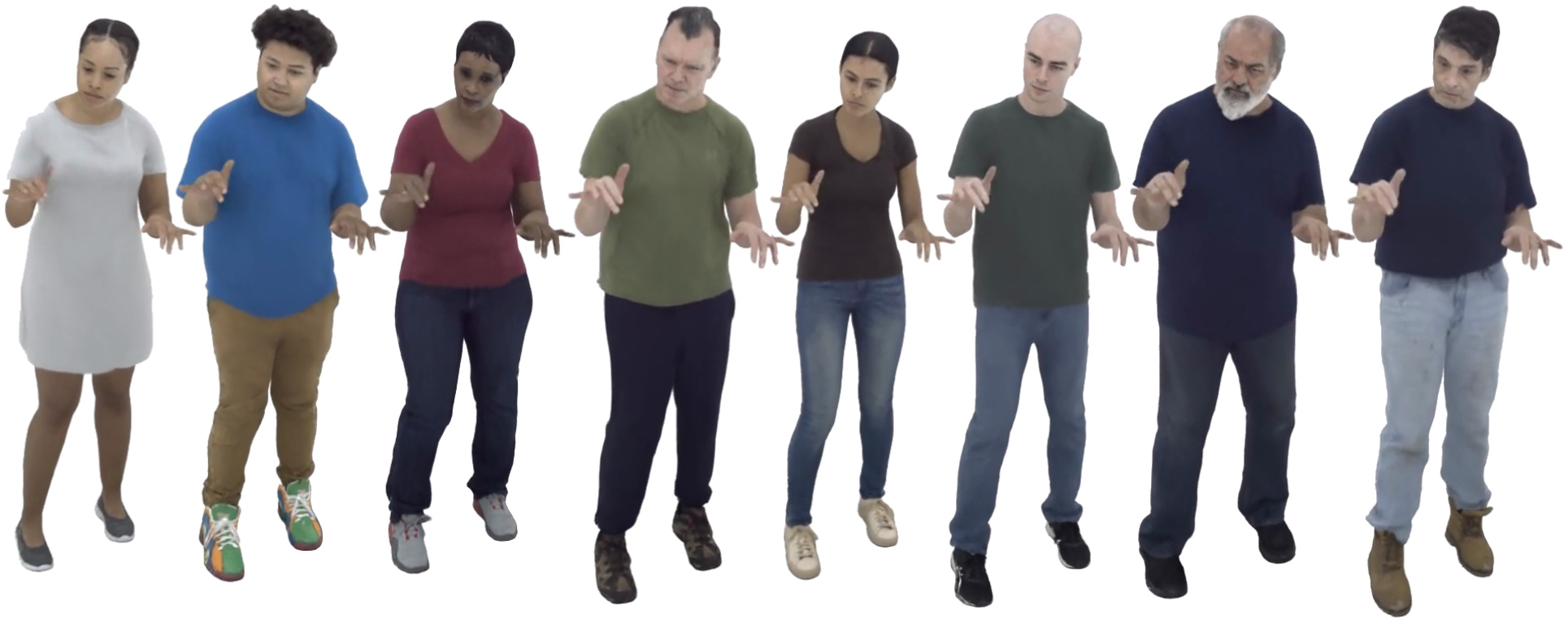}
    \vspace{-0.3cm}
    \caption{\model enables motion transfer showing good generalizability while preserving each avatar's high-quality details.}
    \label{fig:supp-teaser}
\end{center}%
}]

\section{Appendix} 
\begin{figure}[ht!]
    \centering
    \setlength{\unitlength}{0.1\textwidth}
    \begin{picture}(5, 3)
    \put(-0.1, 0){\includegraphics[width=0.5\textwidth]{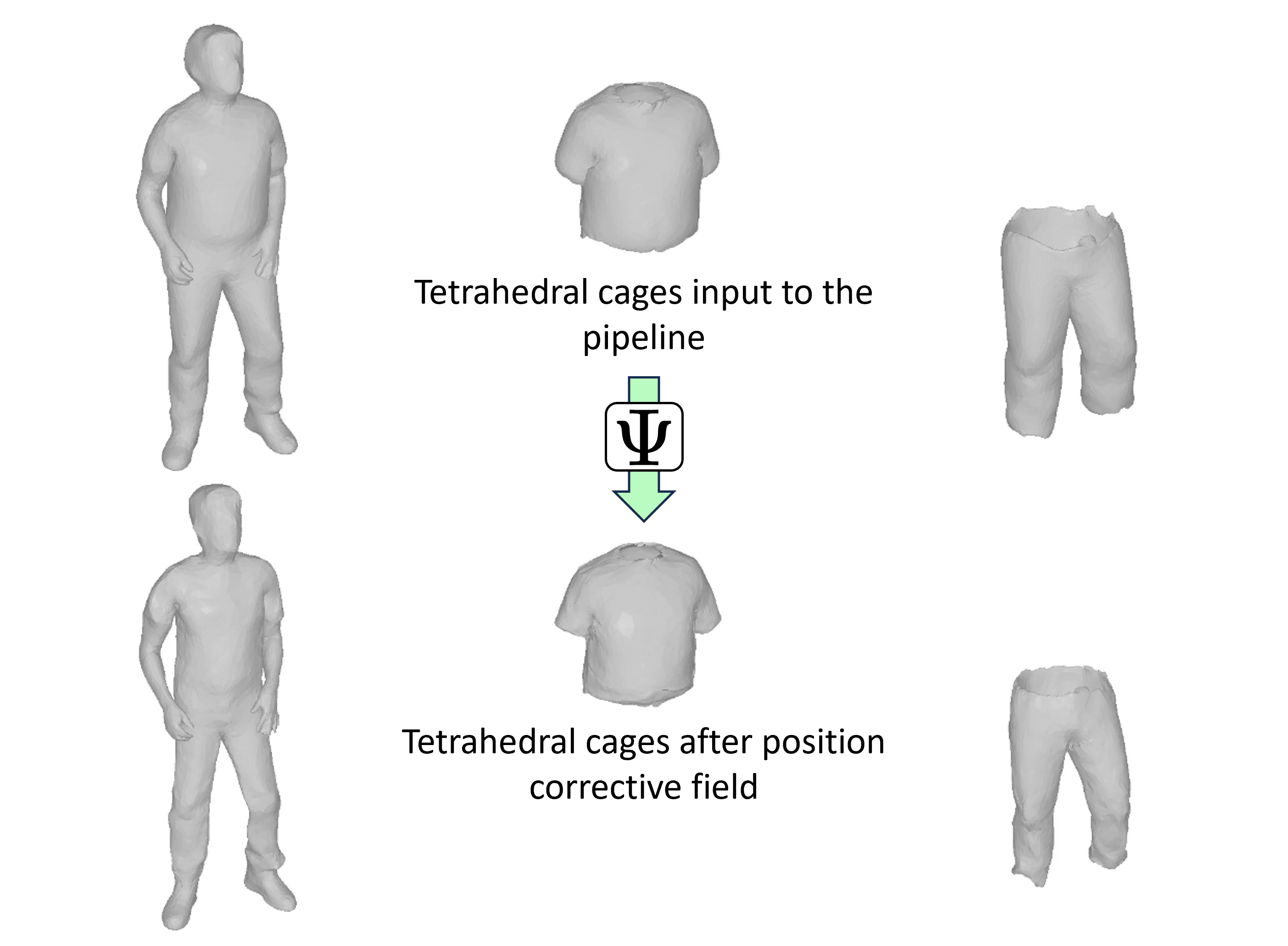}}
    \end{picture}
    \caption{The effect of geometry corrective $\Psi$ field shown on the input tetrahedral meshes before and after the pose corrective field is applied to the vertices.}
    \label{fig:geo_transform}
\end{figure}

This supplemental document presents additional results of our method in the context of garment decomposition and the effect of $\mathcal{L}_{Neo}$ on geometry, qualitative evaluation of garment loss $\mathcal{L}_{Garment}$. Moreover, we show the effect of the corrective field $\Psi$ applied to the input tetrahedrons presented in Figure \ref{fig:geo_transform}. Finally, we present more information about the deformation gradient, color network ablation in the context of shadows and additional comparison to some NeRF-based models like NPC by Su et al. \cite{Su2023NPCNP}.

\begin{figure}[h]
    \centering
    \includegraphics[width=0.8\columnwidth]{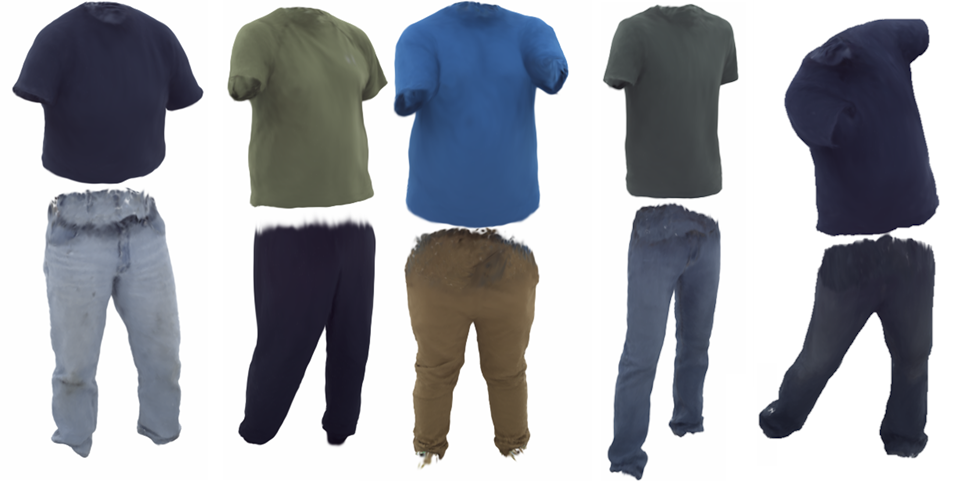}
    \caption{As each component of the avatar is modeled independently, it becomes straightforward to break down the avatar into individual layers. In this demonstration, we showcase the upper and lower segments of the garment. However, it's important to note that we are not confined solely to the garment, enabling us to configure the initial layers in any desired arrangement.}
    \label{fig:garment}
    \vspace{-0.1cm}
\end{figure}

\paragraph{Compositionality:}
One of the important features of our architecture is its composition properties.
We can arbitrarily decompose a given avatar to give segments of interest.
Each of the given segments can undergo different specialized conditioning, for instance, expression codes or keypoints for face or motion vectors for face.
Figure \ref{fig:garment} shows decomposed garments for five different avatars.
Each garment part is independent and can be controlled separately.

\begin{figure}[t!]
\centering
\setlength{\unitlength}{0.1\textwidth}
\begin{picture}(3, 3)
\put(-0.4, 0){\includegraphics[width=0.4\textwidth]{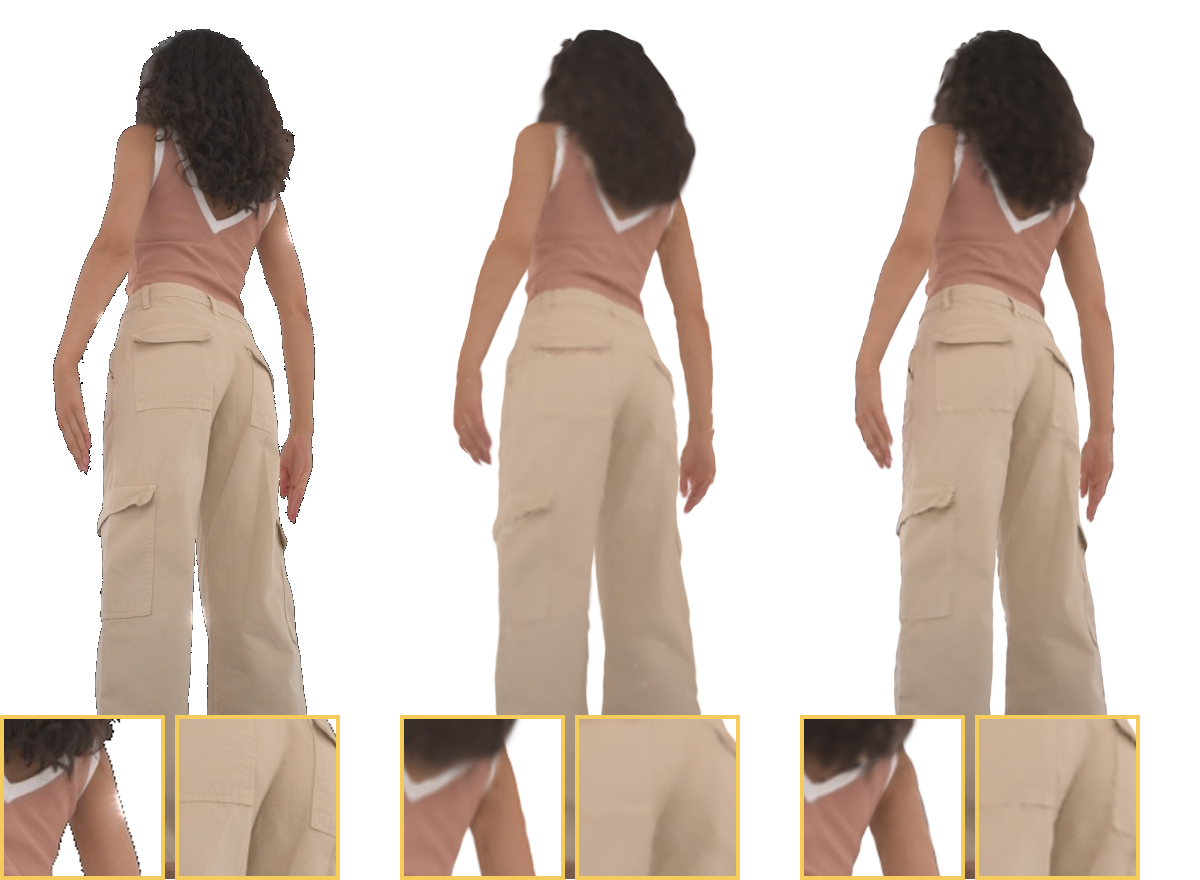}}
\end{picture}
\vspace{0.1cm}
\caption{Our color network replaces the Spherical Harmonics used in the 3DGS with a more compact view-dependent neural network. Here we present the effect of view and pose conditioning on the shadows modeling.}
\label{fig:shadows}
\end{figure}

\paragraph{Regularization Effects:}
We introduced $\mathcal{L}_{Neo}$ to avoid geometry artifacts that could potentially misplace the Gaussians. 
It prevents tetrahedra from losing too much volume, flipping, or diverging in size from the canonical shape. 
Optimization of layered garments will naturally struggle for regions that are either permanently or temporarily covered, resulting in geometric artifacts, which can be alleviated by $\mathcal{L}_{Neo}$ regularization (See Supp. mat for more details).

\begin{figure}[ht!]
    \centering
    \vspace{-1.2cm}
    \setlength{\unitlength}{0.1\textwidth}
    \begin{picture}(3, 3)
    \put(0.5, 0){\includegraphics[width=0.4\columnwidth]{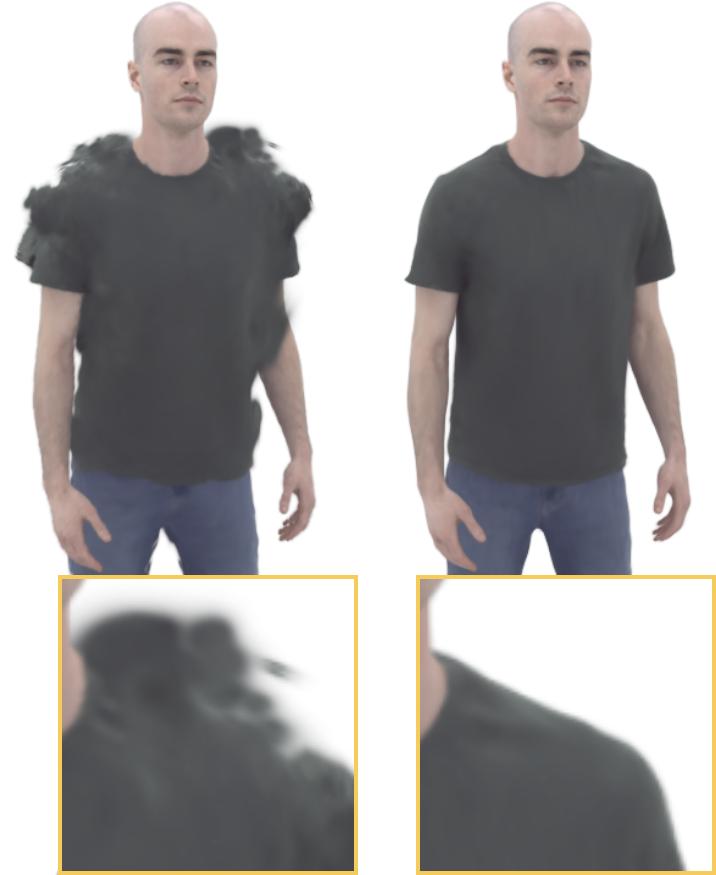}}
    \put(0.4, -0.34){w/o correctives}
    \put(1.8, -0.34){Ours}
    \end{picture}
    \vspace{0.4cm}
    \caption{Effect of corrective networks. Here we disabled corrective networks and only deformed body triangular mesh with LBS.}
    \label{fig:cage_lbs}
\end{figure}

In Figure \ref{fig:cage_lbs}, we show additional ablation of the regularization effect of cage usage. As can be seen, the avatar optimized using only LBS can exhibit artifacts due to incorrect 3D Gaussian orientation during the test time. Using tetrahedral cages and MLP-based correctives can improve their orientation significantly removing the artifacts.

\begin{figure*}[ht!]
    \centering
    \setlength{\unitlength}{0.1\textwidth}
    \begin{picture}(10, 2.5)
    \put(0, 0){\includegraphics[width=1.0\textwidth]{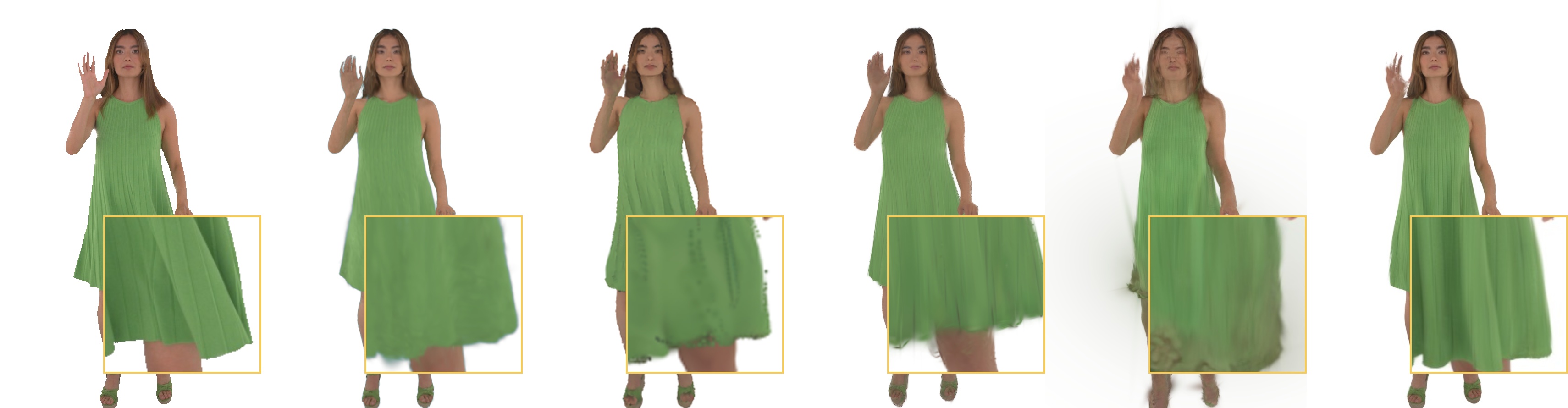}}
    \put(0.4, -0.3){Ground Truth}
    \put(2.4, -0.3){Ours}
    \put(3.6, -0.3){3DGS-Avatar \cite{qian20233dgsavatar}}
    \put(5.3, -0.3){Gaussian Avatar \cite{hu2024gaussianavatar}}
    \put(7.2, -0.3){AG (w/o \textbf{T}) \cite{Li2023AnimatableGL}}
    \put(8.9, -0.3){AG (w/ \textbf{T}) \cite{Li2023AnimatableGL}}
    \end{picture}
    \vspace{0.2cm}
    \caption{For pose-conditioned methods, common failure cases occur with subjects wearing long garments. Methods that do not use a specialized garment template (\textbf{T}), to which primitives are attached, often fail in these scenarios. Although Animitable Gaussians (\textbf{AG}) \cite{Li2023AnimatableGL} achieves the best results when using such a template (/w \textbf{T}), it fails completely without one (w/o \textbf{T}). On the other hand, MLP-based methods are more stable, even when using only the SMPL average body mesh as the underlying template geometry.}
    \label{fig:failure}
\end{figure*}

\paragraph{Failure Cases:}
Human body avatar methods that rely solely on sparse signals, such as joint angle vectors, often struggle to accurately model more complex garment deformations that are independent of the body. Figure \ref{fig:failure} illustrates the most common failure in modeling long garments. All MLP-based solutions use a coarse SMPL mesh to model the avatar. To achieve high-quality results, Animatable GL (AG) \cite{Li2023AnimatableGL} requires a specialized template of the garment tracked per frame. In contrast, MLP-based solutions achieve more stable, albeit incorrect, results using only the SMPL mesh as guiding geometry.

\begin{figure}[h]
    \centering
    \setlength{\unitlength}{0.1\textwidth}
    \begin{picture}(4, 2.4)
    \put(0, 0){\includegraphics[width=0.4\textwidth]{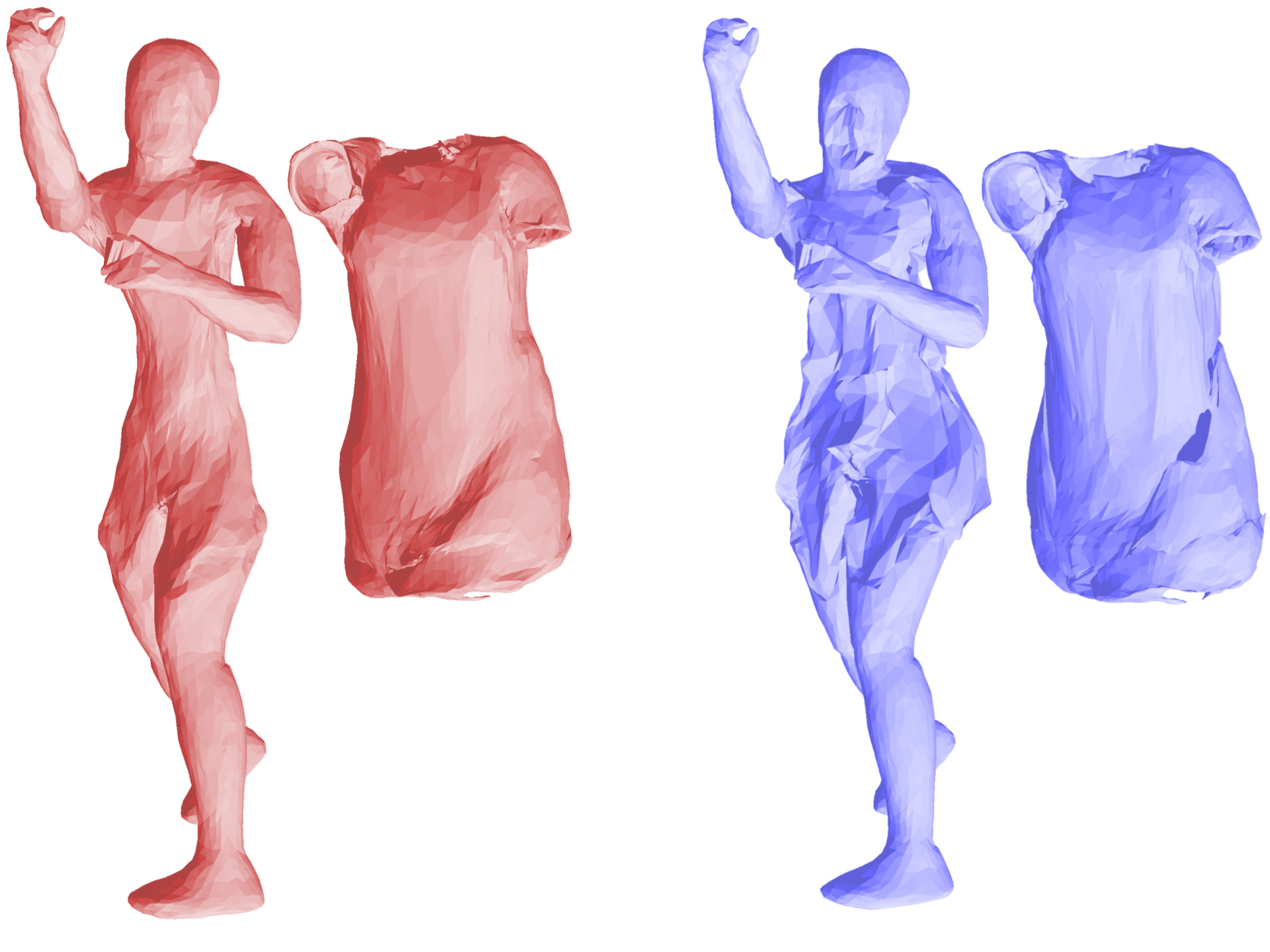}}
    \put(0.8, 0.9){{\small \color{lbsred} w/ $\mathcal{L}_{Neo}$}}
    \put(3.0, 0.9){{\small \color{ffdblue} w/o $\mathcal{L}_{Neo}$}}
    \end{picture}
    \caption{The effect of the tetrahedra regularization $\mathcal{L}_{Neo}$ is mostly visible in the regions which lack supervision or undergo sliding, which covers them for most of the time.}
    \label{fig:neo_ablation}
\end{figure}

\paragraph{Neo-Hookean Term:}
Figure \ref{fig:neo_ablation} shows the regularization effect of the Neo-Hookean term \cite{Macklin2021ACF} to prevent tetrahedrons from sheering or losing volume, especially in places where supervision is not available, e.g., under the garment. 

In table \ref{tab:render_time} we additionally measured the relation between the number of Gaussians, quality, and speed. As can be seen, the best compromise is for 100k and 200k primitives as the tradeoff between speed and quality.

\input{tables/render_time}

\section{Cage Deformation Gradient}

From Sumner et al. \cite{Sumner2004DeformationTF}: Our goal is to encode shape deformation through a differential specification, enabling us to create and use an algorithm that transfers differential changes. Continuum mechanics, which addresses the behavior of materials under external forces \cite{Lai1993}, offers established methods for representing large deformations of solids under load. The deformation gradient, a key concept in this field, provides the exact representation we require. 
\begin{equation}
\tilde{\mathbf{p}} = \mathbf{U}(\mathbf{p}) = 
\begin{bmatrix}
U_1(p_1, p_2, p_3) \\
U_2(p_1, p_2, p_3) \\
U_3(p_1, p_2, p_3)
\end{bmatrix}
\end{equation}
The deformation of an infinitesimal vector \(\mathbf{dp}\) within the solid is determined by the deformation gradient \(\frac{\partial \mathbf{U}}{\partial \mathbf{p}}\). Since \(\mathbf{U}\) maps from \(\mathbb{R}^3\) to \(\mathbb{R}^3\) and varies with position, its gradient is a second-order tensor field:

\begin{equation}
\frac{\partial \mathbf{U}}{\partial \mathbf{p}} = 
\begin{bmatrix}
\frac{\partial U_1}{\partial p_1} & \frac{\partial U_1}{\partial p_2} & \frac{\partial U_1}{\partial p_3} \\
\frac{\partial U_2}{\partial p_1} & \frac{\partial U_2}{\partial p_2} & \frac{\partial U_2}{\partial p_3} \\
\frac{\partial U_3}{\partial p_1} & \frac{\partial U_3}{\partial p_2} & \frac{\partial U_3}{\partial p_3}
\end{bmatrix}
\end{equation}

However, in a more general case, we need to use an approximation of the Jacobian \(\mathbf{U}\) via discretization by triangulation or tetrahedralization for a given shape. In our method, when using cages, we have four vertices \(\{i_1, i_2, i_3, i_4\}\) of a cage for which the deformation gradient can be defined as:

\begin{equation}
\begin{aligned}
\mathbf{J}_j (\mathbf{v}_{i_2} - \mathbf{v}_{i_1}) = \tilde{\mathbf{v}}_{i_2} - \tilde{\mathbf{v}}_{i_1} \\
\mathbf{J}_j (\mathbf{v}_{i_3} - \mathbf{v}_{i_1}) = \tilde{\mathbf{v}}_{i_3} - \tilde{\mathbf{v}}_{i_1} \\
\mathbf{J}_j (\mathbf{v}_{i_4} - \mathbf{v}_{i_1}) = \tilde{\mathbf{v}}_{i_4} - \tilde{\mathbf{v}}_{i_1} \\
\end{aligned}
\end{equation}

\noindent
which in the matrix form equals:

\begin{equation}
\begin{aligned}
\mathbf{J}_j \mathbf{V}_j = \tilde{\mathbf{V}}_j \\
\mathbf{J}_j = \mathbf{V}_j \tilde{\mathbf{V}}_j^{-1} \\
\end{aligned}
\end{equation}

\noindent
where $\mathbf{V}_j$ and $\mathbf{V}_j$ the \(3 \times 3\) matrices and $\mathbf{J}_j$ if the deformation gradient applied to the kernels of each Gaussian primitive $j$ which encapsulates change between canonical and deformed tetrahedrons.

\section{Broader Impact}
Our project focuses on reconstructing a high-fidelity human body avatar from multiview videos, with the capability to extrapolate to poses not originally captured. While our technology is primarily intended for constructive purposes, such as enhancing telepresence or mixed reality applications, we recognize the potential risks of its misuse. Hence, we advocate for advancements in digital media forensics \cite{Rssler2019FaceForensicsLT, Rssler2018FaceForensicsAL} to aid in detecting synthetic media. It is important to highlight that all individuals in our dataset have provided written consent for the use and release of their data.

\begin{figure}[ht!]
    \centering
    \setlength{\unitlength}{0.1\textwidth}
    \begin{picture}(3, 2)
    \put(-1.0, 0){\includegraphics[width=0.5\textwidth]{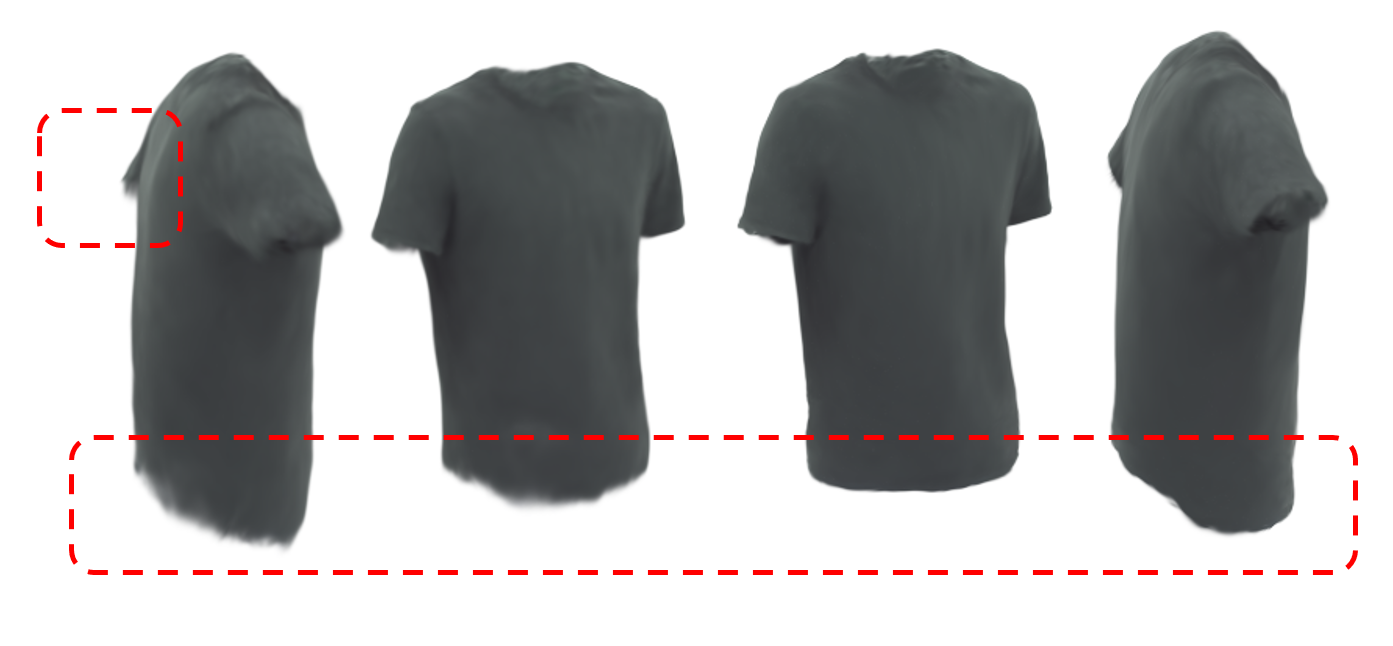}}
    \put(0, -0.2){w/o $\mathcal{L}_{Garment}$}
    \put(2, -0.2){w/ $\mathcal{L}_{Garment}$}
    \end{picture}
    \vspace{0.25cm}
    \caption{The additional supervision $\mathcal{L}_{Garment}$ improves the garment's shape by reducing semitransparent effects at the boundary.}
    \label{fig:garment_ablation}
\end{figure}

\begin{figure}[h]
\centering
\setlength{\unitlength}{0.1\textwidth}
\begin{picture}(3, 3.3)
\put(0, 0){\includegraphics[width=0.65\columnwidth]{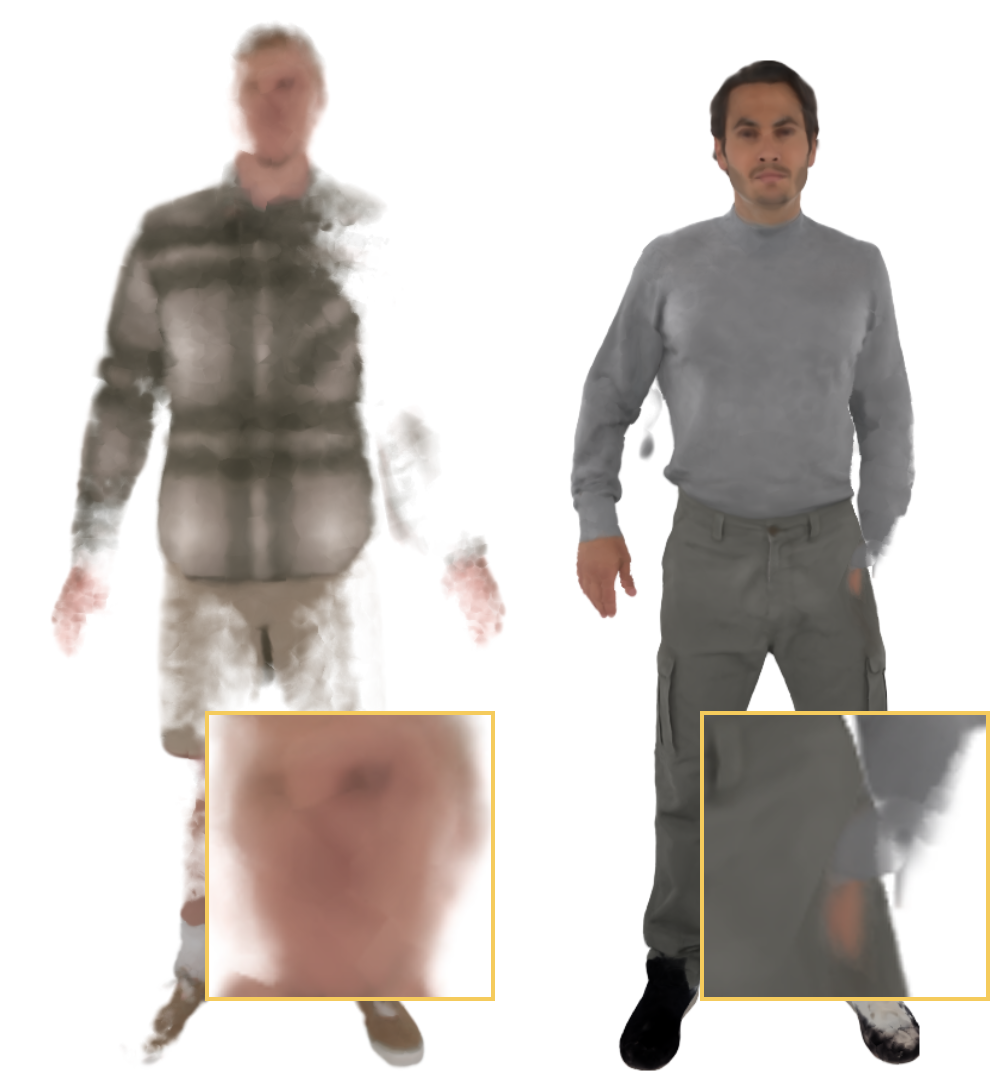}}
\put(0.3, -0.44){{\small 3300 frames}}
\put(2.0, -0.44){{\small 1500 frames}}
\end{picture}
\vspace{0.8cm}
\caption{Results of NPC \cite{Su2023NPCNP} trained with 38 views on short and long dynamic sequences show how a NeRF representation struggles with capturing extended sequences. }
\label{fig:npc}
\end{figure}

As mentioned by Isik et al. \cite{Isik2023TOG} NeRF struggles with capturing long dynamic sequences due to its limited capacity. Evaluation of NPC \cite{Su2023NPCNP} on ActorsHQ sequence shows significant artifacts depicted in Figure \ref{fig:npc}. Moreover, previous generation methods like TAVA \cite{Li2022TAVATA} or ARAH \cite{ARAH:2022:ECCV} are prohibitively slow, especially for high-resolution images like in our case as they were designed to operate on the image with $256 \times 256$ size.

\input{tables/color_error_ablation}

%% file: tables/render_time.tex
\begin{table}[h!]
    \vspace{-0.1cm}
    \centering
    \resizebox{0.8\linewidth}{!}{
        \begin{tabular}{l|rrr|rr}
        Experiment & PSNR $\uparrow$ & LPIPS $\downarrow$ & SSIM $\uparrow$ & FPS $\uparrow$ & $\Delta$FPS \\
        \midrule
        $25k$ Gaussians & \num{29.938485} & \num{0.058495} & \num{0.959905} & 28 & 107\%\\
        $100k$ Gaussians & \num{29.824610} & \num{0.056385} & \num{0.959855} & 26 & 100\%\\
        $200k$ Gaussians & \num{29.863915} & \num{0.056170} & \num{0.960035} & 23 & 88\%\\
        $300k$ Gaussians & \num{29.863735} & \num{0.055635} & \num{0.960010} & 20 & 77\%\\
        \bottomrule
        \end{tabular}
    }
        \caption{Average frame rate per second at $1024\times667$ resolution w.r.t to the amount of Gaussian measured on a Nvidia V100 GPU. 100k Gaussians provide the best rendering-time-to-quality ratio.
    }
    \vspace{-0.1cm}
    \label{tab:render_time}
\end{table}

%% file: tables/color_error_ablation.tex
\begin{table}[h!]
    \vspace{-0.1cm}
    \centering
    \resizebox{0.6\linewidth}{!}{
    \begin{tabular}{l|rrrr}
    Experiment & PSNR $\uparrow$ & LPIPS $\downarrow$ & SSIM $\uparrow$ \\
    \midrule
    Ours & \num{29.824610} & \num{0.056385} & \num{0.959855} \\
    w/o $\mathcal{L}_{Garment}$ & \num{30.139960} & \num{0.057120} & \num{0.960880} \\
    Single layer & \num{29.740205} & \num{0.057435} & \num{0.959365} \\
    \bottomrule
    \end{tabular}
    }
    \caption{Evaluation on our dataset; single-layer avatars incorrectly model sliding garments and garment loss improves the separation of the layers.}
    \label{tab:color_errors_ablation}
    \vspace{-0.1cm}
\end{table}

%% file: main.bbl
\begin{thebibliography}{87}
\providecommand{\natexlab}[1]{#1}
\providecommand{\url}[1]{\texttt{#1}}
\expandafter\ifx\csname urlstyle\endcsname\relax
  \providecommand{\doi}[1]{doi: #1}\else
  \providecommand{\doi}{doi: \begingroup \urlstyle{rm}\Url}\fi

\bibitem[Bagautdinov et~al.(2021)Bagautdinov, Wu, Simon, Prada, Shiratori, Wei, Xu, Sheikh, and Saragih]{Bagautdinov2021DrivingsignalAF}
Timur~M. Bagautdinov, Chenglei Wu, Tomas Simon, Fabi{\'a}n Prada, Takaaki Shiratori, Shih-En Wei, Weipeng Xu, Yaser Sheikh, and Jason~M. Saragih.
\newblock Driving-signal aware full-body avatars.
\newblock \emph{ACM Transactions on Graphics (TOG)}, 40:\penalty0 1 -- 17, 2021.

\bibitem[Cao and Johnson(2023)]{Cao2023HexPlane}
Ang Cao and Justin Johnson.
\newblock Hexplane: A fast representation for dynamic scenes.
\newblock \emph{Proceedings of the IEEE/CVF Conference on Computer Vision and Pattern Recognition (CVPR)}, 2023.

\bibitem[{Cao} et~al.(2019){Cao}, {Hidalgo Martinez}, {Simon}, {Wei}, and {Sheikh}]{OpenPose}
Z. {Cao}, G. {Hidalgo Martinez}, T. {Simon}, S. {Wei}, and Y.~A. {Sheikh}.
\newblock Openpose: Realtime multi-person 2d pose estimation using part affinity fields.
\newblock \emph{IEEE Transactions on Pattern Analysis and Machine Intelligence}, 2019.

\bibitem[Chen et~al.(2023)Chen, Diaz, and Yuksel]{Chen2023ShortestPT}
Heng Chen, Elier Diaz, and Cem Yuksel.
\newblock Shortest path to boundary for self-intersecting meshes.
\newblock \emph{ACM Transactions on Graphics (TOG)}, 42:\penalty0 1 -- 15, 2023.

\bibitem[Fang et~al.(2022)Fang, Yi, Wang, Xie, Zhang, Liu, Nie\ss{}ner, and Tian]{TiNeuVox}
Jiemin Fang, Taoran Yi, Xinggang Wang, Lingxi Xie, Xiaopeng Zhang, Wenyu Liu, Matthias Nie\ss{}ner, and Qi Tian.
\newblock Fast dynamic radiance fields with time-aware neural voxels.
\newblock In \emph{SIGGRAPH Asia 2022 Conference Papers}, 2022.

\bibitem[Feng et~al.(2022)Feng, Yang, Pollefeys, Black, and Bolkart]{Feng2022CapturingAA}
Yao Feng, Jinlong Yang, Marc Pollefeys, Michael~J. Black, and Timo Bolkart.
\newblock Capturing and animation of body and clothing from monocular video.
\newblock \emph{SIGGRAPH Asia 2022 Conference Papers}, 2022.

\bibitem[Feng et~al.(2023)Feng, Liu, Bolkart, Yang, Pollefeys, and Black]{Feng2023DELTA}
Yao Feng, Weiyang Liu, Timo Bolkart, Jinlong Yang, Marc Pollefeys, and Michael~J. Black.
\newblock Learning disentangled avatars with hybrid 3d representations.
\newblock \emph{arXiv}, 2023.

\bibitem[Feng et~al.(2024)Feng, Feng, Shang, Jiang, Yu, Zong, Shao, Wu, Zhou, Jiang, and Yang]{Feng2024GaussianSD}
Yutao Feng, Xiang Feng, Yintong Shang, Ying Jiang, Chang Yu, Zeshun Zong, Tianjia Shao, Hongzhi Wu, Kun Zhou, Chenfanfu Jiang, and Yin Yang.
\newblock Gaussian splashing: Dynamic fluid synthesis with gaussian splatting.
\newblock \emph{ArXiv}, abs/2401.15318, 2024.

\bibitem[Gafni et~al.(2020)Gafni, Thies, Zollhofer, and Nie{\ss}ner]{Gafni2020DynamicNR}
Guy Gafni, Justus Thies, Michael Zollhofer, and Matthias Nie{\ss}ner.
\newblock Dynamic neural radiance fields for monocular 4d facial avatar reconstruction.
\newblock \emph{2021 IEEE/CVF Conference on Computer Vision and Pattern Recognition (CVPR)}, pages 8645--8654, 2020.

\bibitem[Garbin et~al.(2022)Garbin, Kowalski, Estellers, Szymanowicz, Rezaeifar, Shen, Johnson, and Valentin]{Garbin2022Voltemorph}
Stephan~J. Garbin, Marek Kowalski, Virginia Estellers, Stanislaw Szymanowicz, Shideh Rezaeifar, Jingjing Shen, Matthew Johnson, and Julien Valentin.
\newblock Voltemorph: Realtime, controllable and generalisable animation of volumetric representations.
\newblock \emph{CoRR}, abs/2208.00949, 2022.

\bibitem[Hu et~al.(2024)Hu, Zhang, Zhang, Zhou, Liu, Zhang, and Nie]{hu2024gaussianavatar}
Liangxiao Hu, Hongwen Zhang, Yuxiang Zhang, Boyao Zhou, Boning Liu, Shengping Zhang, and Liqiang Nie.
\newblock Gaussianavatar: Towards realistic human avatar modeling from a single video via animatable 3d gaussians.
\newblock In \emph{IEEE/CVF Conference on Computer Vision and Pattern Recognition (CVPR)}, 2024.

\bibitem[Huang et~al.(2006)Huang, Shi, Liu, Zhou, Wei, Teng, Bao, Guo, and Shum]{Huang2006SubspaceGD}
Jin Huang, Xiaohan Shi, Xinguo Liu, Kun Zhou, Li-Yi Wei, Shang-Hua Teng, Hujun Bao, Baining Guo, and Harry Shum.
\newblock Subspace gradient domain mesh deformation.
\newblock \emph{ACM SIGGRAPH 2006 Papers}, 2006.

\bibitem[Isik et~al.(2023)Isik, R{\"{u}}nz, Georgopoulos, Khakhulin, Starck, Agapito, and Nie{\ss}ner]{Isik2023TOG}
Mustafa Isik, Martin R{\"{u}}nz, Markos Georgopoulos, Taras Khakhulin, Jonathan Starck, Lourdes Agapito, and Matthias Nie{\ss}ner.
\newblock Humanrf: High-fidelity neural radiance fields for humans in motion.
\newblock \emph{{ACM} Trans. Graph.}, 42\penalty0 (4):\penalty0 160:1--160:12, 2023.

\bibitem[Jacobson et~al.(2011)Jacobson, Baran, Popovi{\'c}, and Sorkine-Hornung]{Jacobson2011BoundedBW}
Alec Jacobson, Ilya Baran, Jovan Popovi{\'c}, and Olga Sorkine-Hornung.
\newblock Bounded biharmonic weights for real-time deformation.
\newblock \emph{ACM SIGGRAPH 2011 papers}, 2011.

\bibitem[Jiang et~al.(2024)Jiang, Yu, Xie, Li, Feng, Wang, Li, Lau, Gao, Yang, and Jiang]{Jiang2024VRGSAP}
Ying Jiang, Chang Yu, Tianyi Xie, Xuan Li, Yutao Feng, Huamin Wang, Minchen Li, Henry Lau, Feng Gao, Yin Yang, and Chenfanfu Jiang.
\newblock Vr-gs: A physical dynamics-aware interactive gaussian splatting system in virtual reality.
\newblock \emph{ArXiv}, abs/2401.16663, 2024.

\bibitem[Joshi et~al.(2007)Joshi, Meyer, DeRose, Green, and Sanocki]{Joshi2007TOG}
Pushkar Joshi, Mark Meyer, Tony DeRose, Brian Green, and Tom Sanocki.
\newblock Harmonic coordinates for character articulation.
\newblock \emph{{ACM} Trans. Graph.}, 26\penalty0 (3):\penalty0 71, 2007.

\bibitem[Ju et~al.(2005)Ju, Schaefer, and Warren]{Ju2005MeanVC}
Tao Ju, Scott Schaefer, and Joe~D. Warren.
\newblock Mean value coordinates for closed triangular meshes.
\newblock \emph{ACM SIGGRAPH 2005 Papers}, 2005.

\bibitem[Kajiya(1986)]{Kajiya1986TheRE}
James~T. Kajiya.
\newblock The rendering equation.
\newblock \emph{Proceedings of the 13th annual conference on Computer graphics and interactive techniques}, 1986.

\bibitem[Kerbl et~al.(2023)Kerbl, Kopanas, Leimkuehler, and Drettakis]{Kerbl20233DGS}
Bernhard Kerbl, Georgios Kopanas, Thomas Leimkuehler, and George Drettakis.
\newblock 3d gaussian splatting for real-time radiance field rendering.
\newblock \emph{ACM Transactions on Graphics (TOG)}, 42:\penalty0 1 -- 14, 2023.

\bibitem[Kingma and Ba(2014)]{Kingma2014AdamAM}
Diederik~P. Kingma and Jimmy Ba.
\newblock Adam: A method for stochastic optimization.
\newblock \emph{CoRR}, abs/1412.6980, 2014.

\bibitem[Kirillov et~al.(2020)Kirillov, Wu, He, and Girshick]{Kirillov2020PointRend}
Alexander Kirillov, Yuxin Wu, Kaiming He, and Ross~B. Girshick.
\newblock Pointrend: Image segmentation as rendering.
\newblock In \emph{2020 {IEEE/CVF} Conference on Computer Vision and Pattern Recognition, {CVPR} 2020, Seattle, WA, USA, June 13-19, 2020}, pages 9796--9805. Computer Vision Foundation / {IEEE}, 2020.

\bibitem[Kopanas et~al.(2021)Kopanas, Philip, Leimk{\"u}hler, and Drettakis]{Kopanas2021PointBasedNR}
Georgios Kopanas, Julien Philip, Thomas Leimk{\"u}hler, and George Drettakis.
\newblock Point‐based neural rendering with per‐view optimization.
\newblock \emph{Computer Graphics Forum}, 40, 2021.

\bibitem[Lai et~al.(1993)Lai, Rubin, and Krempl]{Lai1993}
Wen~C. Lai, David Rubin, and Erhard Krempl.
\newblock \emph{Introduction to Continuum Mechanics}.
\newblock Pergamon Press, 3rd edition, 1993.

\bibitem[Lassner and Zollh{\"o}fer(2021)]{Lassner2021PulsarES}
Christoph Lassner and Michael Zollh{\"o}fer.
\newblock Pulsar: Efficient sphere-based neural rendering.
\newblock \emph{2021 IEEE/CVF Conference on Computer Vision and Pattern Recognition (CVPR)}, pages 1440--1449, 2021.

\bibitem[Li et~al.(2022{\natexlab{a}})Li, Tanke, Vo, Zollhofer, Gall, Kanazawa, and Lassner]{Li2022TAVATA}
Ruilong Li, Julian Tanke, Minh Vo, Michael Zollhofer, Jurgen Gall, Angjoo Kanazawa, and Christoph Lassner.
\newblock Tava: Template-free animatable volumetric actors.
\newblock \emph{ArXiv}, abs/2206.08929, 2022{\natexlab{a}}.

\bibitem[Li et~al.(2022{\natexlab{b}})Li, Slavcheva, Zollhoefer, Green, Lassner, Kim, Schmidt, Lovegrove, Goesele, Newcombe, and Lv]{li2022neural3dvideosynthesis}
Tianye Li, Mira Slavcheva, Michael Zollhoefer, Simon Green, Christoph Lassner, Changil Kim, Tanner Schmidt, Steven Lovegrove, Michael Goesele, Richard Newcombe, and Zhaoyang Lv.
\newblock Neural 3d video synthesis from multi-view video.
\newblock \emph{Proceedings of the IEEE/CVF Conference on Computer Vision and Pattern Recognition (CVPR)}, 2022{\natexlab{b}}.

\bibitem[Li et~al.(2020)Li, Niklaus, Snavely, and Wang]{Li2020NeuralSF}
Zhengqi Li, Simon Niklaus, Noah Snavely, and Oliver Wang.
\newblock Neural scene flow fields for space-time view synthesis of dynamic scenes.
\newblock \emph{2021 IEEE/CVF Conference on Computer Vision and Pattern Recognition (CVPR)}, pages 6494--6504, 2020.

\bibitem[Li et~al.(2023)Li, Zheng, Wang, and Liu]{Li2023AnimatableGL}
Zhe Li, Zerong Zheng, Lizhen Wang, and Yebin Liu.
\newblock Animatable gaussians: Learning pose-dependent gaussian maps for high-fidelity human avatar modeling.
\newblock \emph{ArXiv}, abs/2311.16096, 2023.

\bibitem[Li et~al.(2024)Li, Zheng, Wang, and Liu]{li2024animatablegaussians}
Zhe Li, Zerong Zheng, Lizhen Wang, and Yebin Liu.
\newblock Animatable gaussians: Learning pose-dependent gaussian maps for high-fidelity human avatar modeling.
\newblock In \emph{Proceedings of the IEEE/CVF Conference on Computer Vision and Pattern Recognition (CVPR)}, 2024.

\bibitem[Liu et~al.(2021)Liu, Habermann, Rudnev, Sarkar, Gu, and Theobalt]{Liu2021NeuralA}
Lingjie Liu, Marc Habermann, Viktor Rudnev, Kripasindhu Sarkar, Jiatao Gu, and Christian Theobalt.
\newblock Neural actor.
\newblock \emph{ACM Transactions on Graphics (TOG)}, 40:\penalty0 1 -- 16, 2021.

\bibitem[Lombardi et~al.(2021)Lombardi, Simon, Schwartz, Zollhoefer, Sheikh, and Saragih]{Lombardi2021MixtureOV}
Stephen Lombardi, Tomas Simon, Gabriel Schwartz, Michael Zollhoefer, Yaser Sheikh, and Jason~M. Saragih.
\newblock Mixture of volumetric primitives for efficient neural rendering.
\newblock \emph{ACM Transactions on Graphics (TOG)}, 40:\penalty0 1 -- 13, 2021.

\bibitem[Loper et~al.(2015)Loper, Mahmood, Romero, Pons-Moll, and Black]{Loper2015SMPLAS}
Matthew Loper, Naureen Mahmood, Javier Romero, Gerard Pons-Moll, and Michael~J. Black.
\newblock Smpl: A skinned multi-person linear model.
\newblock \emph{Seminal Graphics Papers: Pushing the Boundaries, Volume 2}, 2015.

\bibitem[Lorensen and Cline(1987)]{Lorensen1987MarchingCubes}
William~E. Lorensen and Harvey~E. Cline.
\newblock Marching cubes: A high resolution 3d surface construction algorithm.
\newblock In \emph{SIGGRAPH '87: Proceedings of the 14th Annual Conference on Computer Graphics and Interactive Techniques}, pages 163--169, 1987.

\bibitem[Luo et~al.(2024)Luo, Min, Zhao, Jiang, Zhang, Zhang, Yang, Xu, and Yu]{Luo2024GaussianHairHM}
Haimin Luo, Ouyang Min, Zijun Zhao, Suyi Jiang, Longwen Zhang, Qixuan Zhang, Wei Yang, Lan Xu, and Jingyi Yu.
\newblock Gaussianhair: Hair modeling and rendering with light-aware gaussians.
\newblock \emph{ArXiv}, abs/2402.10483, 2024.

\bibitem[Ma et~al.(2021)Ma, Yang, Tang, and Black]{Ma2021ThePO}
Qianli Ma, Jinlong Yang, Siyu Tang, and Michael~J. Black.
\newblock The power of points for modeling humans in clothing.
\newblock \emph{2021 IEEE/CVF International Conference on Computer Vision (ICCV)}, pages 10954--10964, 2021.

\bibitem[Macklin and M{\"u}ller(2021)]{Macklin2021ACF}
Miles Macklin and Matthias M{\"u}ller.
\newblock A constraint-based formulation of stable neo-hookean materials.
\newblock \emph{Proceedings of the 14th ACM SIGGRAPH Conference on Motion, Interaction and Games}, 2021.

\bibitem[Martin-Brualla et~al.(2020)Martin-Brualla, Radwan, Sajjadi, Barron, Dosovitskiy, and Duckworth]{MartinBrualla2020NeRFIT}
Ricardo Martin-Brualla, Noha Radwan, Mehdi S.~M. Sajjadi, Jonathan~T. Barron, Alexey Dosovitskiy, and Daniel Duckworth.
\newblock Nerf in the wild: Neural radiance fields for unconstrained photo collections.
\newblock \emph{2021 IEEE/CVF Conference on Computer Vision and Pattern Recognition (CVPR)}, pages 7206--7215, 2020.

\bibitem[Martinez et~al.(2024)Martinez, Kim, Romero, Bagautdinov, Saito, Yu, Anderson, Zollhöfer, Wang, Bai, Li, Wei, Joshi, Borsos, Simon, Saragih, Theodosis, Greene, Josyula, Maeta, Jewett, Venshtain, Heilman, Chen, Fu, Elshaer, Du, Wu, Chen, Kang, Wu, Emad, Longay, Brewer, Shah, Booth, Koska, Haidle, Andromalos, Hsu, Dauer, Selednik, Godisart, Ardisson, Cipperly, Humberston, Farr, Hansen, Guo, Braun, Krenn, Wen, Evans, Fadeeva, Stewart, Schwartz, Gupta, Moon, Guo, Dong, Xu, Shiratori, Prada, Pires, Peng, Buffalini, Trimble, McPhail, Schoeller, and Sheikh]{martinez2024codec}
Julieta Martinez, Emily Kim, Javier Romero, Timur Bagautdinov, Shunsuke Saito, Shoou-I Yu, Stuart Anderson, Michael Zollhöfer, Te-Li Wang, Shaojie Bai, Chenghui Li, Shih-En Wei, Rohan Joshi, Wyatt Borsos, Tomas Simon, Jason Saragih, Paul Theodosis, Alexander Greene, Anjani Josyula, Silvio~Mano Maeta, Andrew~I. Jewett, Simon Venshtain, Christopher Heilman, Yueh-Tung Chen, Sidi Fu, Mohamed Ezzeldin~A. Elshaer, Tingfang Du, Longhua Wu, Shen-Chi Chen, Kai Kang, Michael Wu, Youssef Emad, Steven Longay, Ashley Brewer, Hitesh Shah, James Booth, Taylor Koska, Kayla Haidle, Matt Andromalos, Joanna Hsu, Thomas Dauer, Peter Selednik, Tim Godisart, Scott Ardisson, Matthew Cipperly, Ben Humberston, Lon Farr, Bob Hansen, Peihong Guo, Dave Braun, Steven Krenn, He Wen, Lucas Evans, Natalia Fadeeva, Matthew Stewart, Gabriel Schwartz, Divam Gupta, Gyeongsik Moon, Kaiwen Guo, Yuan Dong, Yichen Xu, Takaaki Shiratori, Fabian Prada, Bernardo~R. Pires, Bo Peng, Julia Buffalini, Autumn Trimble, Kevyn McPhail, Melissa Schoeller, and
  Yaser Sheikh.
\newblock {Codec Avatar Studio: Paired Human Captures for Complete, Driveable, and Generalizable Avatars}.
\newblock \emph{NeurIPS Track on Datasets and Benchmarks}, 2024.

\bibitem[Mihajlovic et~al.(2022)Mihajlovic, Bansal, Zollhoefer, Tang, and Saito]{Mihajlovic2022KeypointNeRFGI}
Marko Mihajlovic, Aayush Bansal, Michael Zollhoefer, Siyu Tang, and Shunsuke Saito.
\newblock Keypointnerf: Generalizing image-based volumetric avatars using relative spatial encoding of keypoints.
\newblock \emph{ArXiv}, abs/2205.04992, 2022.

\bibitem[Mildenhall et~al.(2020)Mildenhall, Srinivasan, Tancik, Barron, Ramamoorthi, and Ng]{Mildenhall2020NeRF}
Ben Mildenhall, Pratul~P. Srinivasan, Matthew Tancik, Jonathan~T. Barron, Ravi Ramamoorthi, and Ren Ng.
\newblock Nerf.
\newblock \emph{Communications of the ACM}, 65:\penalty0 99 -- 106, 2020.

\bibitem[Nieto and Sus{\'\i}n(2012)]{Nieto2012Deformation}
Jes{\'u}s~R Nieto and Antonio Sus{\'\i}n.
\newblock Cage based deformations: a survey.
\newblock In \emph{Deformation Models: Tracking, Animation and Applications}, pages 75--99. Springer, 2012.

\bibitem[Pang et~al.(2023)Pang, Zhu, Kortylewski, Theobalt, and Habermann]{Pang2023ASHAG}
Haokai Pang, Heming Zhu, Adam Kortylewski, Christian Theobalt, and Marc Habermann.
\newblock Ash: Animatable gaussian splats for efficient and photoreal human rendering.
\newblock \emph{ArXiv}, abs/2312.05941, 2023.

\bibitem[Park et~al.(2019)Park, Florence, Straub, Newcombe, and Lovegrove]{Park2019Deepsdf}
Jeong~Joon Park, Peter Florence, Julian Straub, Richard Newcombe, and Steven Lovegrove.
\newblock Deepsdf: Learning continuous signed distance functions for shape representation.
\newblock In \emph{Proceedings of the IEEE/CVF conference on computer vision and pattern recognition}, pages 165--174, 2019.

\bibitem[Park et~al.(2020)Park, Sinha, Barron, Bouaziz, Goldman, Seitz, and Martin-Brualla]{Park2020NerfiesDN}
Keunhong Park, U. Sinha, Jonathan~T. Barron, Sofien Bouaziz, Dan~B. Goldman, Steven~M. Seitz, and Ricardo Martin-Brualla.
\newblock Nerfies: Deformable neural radiance fields.
\newblock \emph{2021 IEEE/CVF International Conference on Computer Vision (ICCV)}, pages 5845--5854, 2020.

\bibitem[Park et~al.(2021)Park, Sinha, Hedman, Barron, Bouaziz, Goldman, Martin-Brualla, and Seitz]{Park2021HyperNeRF}
Keunhong Park, U. Sinha, Peter Hedman, Jonathan~T. Barron, Sofien Bouaziz, Dan~B. Goldman, Ricardo Martin-Brualla, and Steven~M. Seitz.
\newblock Hypernerf.
\newblock \emph{ACM Transactions on Graphics (TOG)}, 40:\penalty0 1 -- 12, 2021.

\bibitem[Peng et~al.(2020)Peng, Zhang, Xu, Wang, Shuai, Bao, and Zhou]{Peng2020NeuralBI}
Sida Peng, Yuanqing Zhang, Yinghao Xu, Qianqian Wang, Qing Shuai, Hujun Bao, and Xiaowei Zhou.
\newblock Neural body: Implicit neural representations with structured latent codes for novel view synthesis of dynamic humans.
\newblock \emph{2021 IEEE/CVF Conference on Computer Vision and Pattern Recognition (CVPR)}, pages 9050--9059, 2020.

\bibitem[Peng et~al.(2022)Peng, Yan, Liu, Cheng, Guan, Pan, Zhai, and Yang]{Peng2022CageNeRF}
Yicong Peng, Yichao Yan, Shengqi Liu, Yuhao Cheng, Shanyan Guan, Bowen Pan, Guangtao Zhai, and Xiaokang Yang.
\newblock Cagenerf: Cage-based neural radiance field for generalized 3d deformation and animation.
\newblock In \emph{NeurIPS}, 2022.

\bibitem[Prinzler et~al.(2022)Prinzler, Hilliges, and Thies]{Prinzler2022DINERDI}
Malte Prinzler, Otmar Hilliges, and Justus Thies.
\newblock Diner: Depth-aware image-based neural radiance fields.
\newblock \emph{2023 IEEE/CVF Conference on Computer Vision and Pattern Recognition (CVPR)}, pages 12449--12459, 2022.

\bibitem[Prokudin et~al.(2023)Prokudin, Ma, Raafat, Valentin, and Tang]{Prokudin2023dynamic}
Sergey Prokudin, Qianli Ma, Maxime Raafat, Julien Valentin, and Siyu Tang.
\newblock Dynamic point fields.
\newblock \emph{arXiv preprint arXiv:2304.02626}, 2023.

\bibitem[Qian et~al.(2023)Qian, Kirschstein, Schoneveld, Davoli, Giebenhain, and Nießner]{qian2023gaussianavatars}
Shenhan Qian, Tobias Kirschstein, Liam Schoneveld, Davide Davoli, Simon Giebenhain, and Matthias Nießner.
\newblock Gaussianavatars: Photorealistic head avatars with rigged 3d gaussians, 2023.

\bibitem[Qian et~al.(2024)Qian, Wang, Mihajlovic, Geiger, and Tang]{qian20233dgsavatar}
Zhiyin Qian, Shaofei Wang, Marko Mihajlovic, Andreas Geiger, and Siyu Tang.
\newblock 3dgs-avatar: Animatable avatars via deformable 3d gaussian splatting.
\newblock 2024.

\bibitem[Ramamoorthi and Hanrahan(2001)]{Ramamoorthi2001AnER}
Ravi Ramamoorthi and Pat Hanrahan.
\newblock An efficient representation for irradiance environment maps.
\newblock \emph{Proceedings of the 28th annual conference on Computer graphics and interactive techniques}, 2001.

\bibitem[Remelli et~al.(2022)Remelli, Bagautdinov, Saito, Wu, Simon, Wei, Guo, Cao, Prada, Saragih, and Sheikh]{Remelli2022DrivableVA}
Edoardo Remelli, Timur~M. Bagautdinov, Shunsuke Saito, Chenglei Wu, Tomas Simon, Shih-En Wei, Kaiwen Guo, Zhe Cao, Fabi{\'a}n Prada, Jason~M. Saragih, and Yaser Sheikh.
\newblock Drivable volumetric avatars using texel-aligned features.
\newblock \emph{ACM SIGGRAPH 2022 Conference Proceedings}, 2022.

\bibitem[R{\"o}ssler et~al.(2018)R{\"o}ssler, Cozzolino, Verdoliva, Riess, Thies, and Nie{\ss}ner]{Rssler2018FaceForensicsAL}
Andreas R{\"o}ssler, Davide Cozzolino, Luisa Verdoliva, Christian Riess, Justus Thies, and Matthias Nie{\ss}ner.
\newblock Faceforensics: A large-scale video dataset for forgery detection in human faces.
\newblock \emph{ArXiv}, abs/1803.09179, 2018.

\bibitem[R{\"o}ssler et~al.(2019)R{\"o}ssler, Cozzolino, Verdoliva, Riess, Thies, and Nie{\ss}ner]{Rssler2019FaceForensicsLT}
Andreas R{\"o}ssler, Davide Cozzolino, Luisa Verdoliva, Christian Riess, Justus Thies, and Matthias Nie{\ss}ner.
\newblock Faceforensics++: Learning to detect manipulated facial images.
\newblock \emph{2019 IEEE/CVF International Conference on Computer Vision (ICCV)}, pages 1--11, 2019.

\bibitem[Saito et~al.(2023)Saito, Schwartz, Simon, Li, and Nam]{Saito2023RelightableGC}
Shunsuke Saito, Gabriel Schwartz, Tomas Simon, Junxuan Li, and Giljoo Nam.
\newblock Relightable gaussian codec avatars.
\newblock \emph{ArXiv}, abs/2312.03704, 2023.

\bibitem[Si(2013)]{Si2013TetGenAQ}
Hang Si.
\newblock Tetgen: A quality tetrahedral mesh generator and a 3d delaunay triangulator (version 1.5 --- user's manual).
\newblock 2013.

\bibitem[Su et~al.(2021)Su, Yu, Zollhoefer, and Rhodin]{Su2021ANeRFAN}
Shih-Yang Su, Frank Yu, Michael Zollhoefer, and Helge Rhodin.
\newblock A-nerf: Articulated neural radiance fields for learning human shape, appearance, and pose.
\newblock In \emph{Neural Information Processing Systems}, 2021.

\bibitem[Su et~al.(2022)Su, Bagautdinov, and Rhodin]{Su2022DANBODA}
Shih-Yang Su, Timur~M. Bagautdinov, and Helge Rhodin.
\newblock Danbo: Disentangled articulated neural body representations via graph neural networks.
\newblock In \emph{European Conference on Computer Vision}, 2022.

\bibitem[Su et~al.(2023)Su, Bagautdinov, and Rhodin]{Su2023NPCNP}
Shih-Yang Su, Timur~M. Bagautdinov, and Helge Rhodin.
\newblock Npc: Neural point characters from video.
\newblock \emph{ArXiv}, abs/2304.02013, 2023.

\bibitem[Sumner and Popovi{\'c}(2004)]{Sumner2004DeformationTF}
Robert~W. Sumner and Jovan Popovi{\'c}.
\newblock Deformation transfer for triangle meshes.
\newblock \emph{ACM SIGGRAPH 2004 Papers}, 2004.

\bibitem[Tan and Le(2019)]{Tan2019EfficientNet}
Mingxing Tan and Quoc~V. Le.
\newblock Efficientnet: Rethinking model scaling for convolutional neural networks.
\newblock In \emph{Proceedings of the 36th International Conference on Machine Learning, {ICML} 2019, 9-15 June 2019, Long Beach, California, {USA}}, pages 6105--6114. {PMLR}, 2019.

\bibitem[Tewari et~al.(2021)Tewari, Fried, Thies, Sitzmann, Lombardi, Xu, Simon, Nie{\ss}ner, Tretschk, Liu, Mildenhall, Srinivasan, Pandey, Orts-Escolano, Fanello, Guo, Wetzstein, y~Zhu, Theobalt, Agrawala, Goldman, and Zollh{\"o}fer]{Tewari2021AdvancesIN}
Ayush Tewari, Otto Fried, Justus Thies, Vincent Sitzmann, S. Lombardi, Z. Xu, Tanaba Simon, Matthias Nie{\ss}ner, Edgar Tretschk, L. Liu, Ben Mildenhall, Pranatharthi Srinivasan, R. Pandey, Sergio Orts-Escolano, S. Fanello, M.~Guang Guo, Gordon Wetzstein, J y Zhu, Christian Theobalt, Manju Agrawala, Donald~B. Goldman, and Michael Zollh{\"o}fer.
\newblock Advances in neural rendering.
\newblock \emph{Computer Graphics Forum}, 41, 2021.

\bibitem[Tewari et~al.(2020)Tewari, Fried, Thies, Sitzmann, Lombardi, Sunkavalli, Martin-Brualla, Simon, Saragih, Nie{\ss}ner, Pandey, Fanello, Wetzstein, Zhu, Theobalt, Agrawala, Shechtman, Goldman, and Zollhofer]{Tewari2020StateOT}
Ayush~Kumar Tewari, Ohad Fried, Justus Thies, Vincent Sitzmann, Stephen Lombardi, Kalyan Sunkavalli, Ricardo Martin-Brualla, Tomas Simon, Jason~M. Saragih, Matthias Nie{\ss}ner, Rohit Pandey, S. Fanello, Gordon Wetzstein, Jun-Yan Zhu, Christian Theobalt, Maneesh Agrawala, Eli Shechtman, Dan~B. Goldman, and Michael Zollhofer.
\newblock State of the art on neural rendering.
\newblock \emph{Computer Graphics Forum}, 39, 2020.

\bibitem[Wang et~al.(2023)Wang, Zhao, Sun, Zhang, Zhang, Yu, and Liu]{wang2023styleavatar}
Lizhen Wang, Xiaochen Zhao, Jingxiang Sun, Yuxiang Zhang, Hongwen Zhang, Tao Yu, and Yebin Liu.
\newblock Styleavatar: Real-time photo-realistic portrait avatar from a single video.
\newblock In \emph{ACM SIGGRAPH 2023 Conference Proceedings}, 2023.

\bibitem[Wang et~al.(2022)Wang, Schwarz, Geiger, and Tang]{ARAH:2022:ECCV}
Shaofei Wang, Katja Schwarz, Andreas Geiger, and Siyu Tang.
\newblock Arah: Animatable volume rendering of articulated human sdfs.
\newblock In \emph{European Conference on Computer Vision}, 2022.

\bibitem[Wang et~al.(2019)Wang, Serena, Wu, {\"O}ztireli, and Sorkine-Hornung]{Wang2019DifferentiableSS}
Yifan Wang, Felice Serena, Shihao Wu, Cengiz {\"O}ztireli, and Olga Sorkine-Hornung.
\newblock Differentiable surface splatting for point-based geometry processing.
\newblock \emph{ACM Transactions on Graphics (TOG)}, 38:\penalty0 1 -- 14, 2019.

\bibitem[Wang et~al.(2020)Wang, Aigerman, Kim, Chaudhuri, and Sorkine{-}Hornung]{Wang2020NeuralCages}
Yifan Wang, Noam Aigerman, Vladimir~G. Kim, Siddhartha Chaudhuri, and Olga Sorkine{-}Hornung.
\newblock Neural cages for detail-preserving 3d deformations.
\newblock In \emph{2020 {IEEE/CVF} Conference on Computer Vision and Pattern Recognition, {CVPR} 2020, Seattle, WA, USA, June 13-19, 2020}, pages 72--80. Computer Vision Foundation / {IEEE}, 2020.

\bibitem[Weng et~al.(2022)Weng, Curless, Srinivasan, Barron, and Kemelmacher{-}Shlizerman]{Weng2022CVPR}
Chung{-}Yi Weng, Brian Curless, Pratul~P. Srinivasan, Jonathan~T. Barron, and Ira Kemelmacher{-}Shlizerman.
\newblock Humannerf: Free-viewpoint rendering of moving people from monocular video.
\newblock In \emph{{IEEE/CVF} Conference on Computer Vision and Pattern Recognition, {CVPR} 2022, New Orleans, LA, USA, June 18-24, 2022}, pages 16189--16199. {IEEE}, 2022.

\bibitem[Wu et~al.(2024)Wu, Yi, Fang, Xie, Zhang, Wei, Liu, Tian, and Xinggang]{wu20234dgaussians}
Guanjun Wu, Taoran Yi, Jiemin Fang, Lingxi Xie, Xiaopeng Zhang, Wei Wei, Wenyu Liu, Qi Tian, and Wang Xinggang.
\newblock 4d gaussian splatting for real-time dynamic scene rendering.
\newblock \emph{Proceedings of the IEEE/CVF Conference on Computer Vision and Pattern Recognition (CVPR)}, 2024.

\bibitem[Xiang et~al.(2021)Xiang, Prada, Bagautdinov, Xu, Dong, Wen, Hodgins, and Wu]{Xiang2021ModelingCA}
Donglai Xiang, Fabi{\'a}n Prada, Timur~M. Bagautdinov, Weipeng Xu, Yuan Dong, He Wen, Jessica~K. Hodgins, and Chenglei Wu.
\newblock Modeling clothing as a separate layer for an animatable human avatar.
\newblock \emph{ACM Transactions on Graphics (TOG)}, 40:\penalty0 1 -- 15, 2021.

\bibitem[Xiang et~al.(2023{\natexlab{a}})Xiang, Prada, Cao, Guo, Wu, Hodgins, and Bagautdinov]{Xiang2023DrivableAC}
Donglai Xiang, Fabi{\'a}n Prada, Zhe Cao, Kaiwen Guo, Chenglei Wu, Jessica~K. Hodgins, and Timur~M. Bagautdinov.
\newblock Drivable avatar clothing: Faithful full-body telepresence with dynamic clothing driven by sparse rgb-d input.
\newblock 2023{\natexlab{a}}.

\bibitem[Xiang et~al.(2023{\natexlab{b}})Xiang, Gao, Guo, and yong Zhang]{Xiang2023FlashAvatarHD}
Jun Xiang, Xuan Gao, Yudong Guo, and Ju yong Zhang.
\newblock Flashavatar: High-fidelity digital avatar rendering at 300fps.
\newblock \emph{ArXiv}, abs/2312.02214, 2023{\natexlab{b}}.

\bibitem[Xie et~al.(2023)Xie, Zong, Qiu, Li, Feng, Yang, and Jiang]{xie2023physgaussian}
Tianyi Xie, Zeshun Zong, Yuxing Qiu, Xuan Li, Yutao Feng, Yin Yang, and Chenfanfu Jiang.
\newblock Physgaussian: Physics-integrated 3d gaussians for generative dynamics, 2023.

\bibitem[Xu et~al.(2022)Xu, Xu, Philip, Bi, Shu, Sunkavalli, and Neumann]{Xu2022PointNeRFPN}
Qiangeng Xu, Zexiang Xu, Julien Philip, Sai Bi, Zhixin Shu, Kalyan Sunkavalli, and Ulrich Neumann.
\newblock Point-nerf: Point-based neural radiance fields.
\newblock \emph{2022 IEEE/CVF Conference on Computer Vision and Pattern Recognition (CVPR)}, pages 5428--5438, 2022.

\bibitem[Xu et~al.(2023)Xu, Chen, Li, Zhang, Wang, Zheng, and Liu]{Xu2023GaussianHA}
Yuelang Xu, Benwang Chen, Zhe Li, Hongwen Zhang, Lizhen Wang, Zerong Zheng, and Yebin Liu.
\newblock Gaussian head avatar: Ultra high-fidelity head avatar via dynamic gaussians.
\newblock \emph{ArXiv}, abs/2312.03029, 2023.

\bibitem[Yang et~al.(2024)Yang, Yang, Pan, and Zhang]{yang2023gs4d}
Zeyu Yang, Hongye Yang, Zijie Pan, and Li Zhang.
\newblock Real-time photorealistic dynamic scene representation and rendering with 4d gaussian splatting.
\newblock In \emph{International Conference on Learning Representations (ICLR)}, 2024.

\bibitem[Zhang et~al.(2023)Zhang, Feng, Kulits, Wen, Thies, and Black]{Zhang2023TextGuidedGA}
Hao Zhang, Yao Feng, Peter Kulits, Yandong Wen, Justus Thies, and Michael~J. Black.
\newblock Text-guided generation and editing of compositional 3d avatars.
\newblock \emph{ArXiv}, abs/2309.07125, 2023.

\bibitem[Zhang et~al.(2018)Zhang, Isola, Efros, Shechtman, and Wang]{Zhang2018TheUE}
Richard Zhang, Phillip Isola, Alexei~A. Efros, Eli Shechtman, and Oliver Wang.
\newblock The unreasonable effectiveness of deep features as a perceptual metric.
\newblock \emph{2018 IEEE/CVF Conference on Computer Vision and Pattern Recognition}, pages 586--595, 2018.

\bibitem[Zheng et~al.(2023)Zheng, Zhou, Shao, Liu, Zhang, Nie, and Liu]{Zheng2023GPSGaussianGP}
Shunyuan Zheng, Boyao Zhou, Ruizhi Shao, Boning Liu, Shengping Zhang, Liqiang Nie, and Yebin Liu.
\newblock Gps-gaussian: Generalizable pixel-wise 3d gaussian splatting for real-time human novel view synthesis.
\newblock \emph{ArXiv}, abs/2312.02155, 2023.

\bibitem[Zheng et~al.(2022{\natexlab{a}})Zheng, Wang, Wetzstein, Black, and Hilliges]{Zheng2022PointAvatarDP}
Yufeng Zheng, Yifan Wang, Gordon Wetzstein, Michael~J. Black, and Otmar Hilliges.
\newblock Pointavatar: Deformable point-based head avatars from videos.
\newblock \emph{2023 IEEE/CVF Conference on Computer Vision and Pattern Recognition (CVPR)}, pages 21057--21067, 2022{\natexlab{a}}.

\bibitem[Zheng et~al.(2022{\natexlab{b}})Zheng, Huang, Yu, Zhang, Guo, and Liu]{Zheng2022StructuredLR}
Zerong Zheng, Han Huang, Tao Yu, Hongwen Zhang, Yandong Guo, and Yebin Liu.
\newblock Structured local radiance fields for human avatar modeling.
\newblock \emph{2022 IEEE/CVF Conference on Computer Vision and Pattern Recognition (CVPR)}, pages 15872--15882, 2022{\natexlab{b}}.

\bibitem[Zielonka et~al.(2022)Zielonka, Bolkart, and Thies]{Zielonka2022InstantVH}
Wojciech Zielonka, Timo Bolkart, and Justus Thies.
\newblock Instant volumetric head avatars.
\newblock \emph{2023 IEEE/CVF Conference on Computer Vision and Pattern Recognition (CVPR)}, pages 4574--4584, 2022.

\bibitem[Zielonka et~al.(2024)Zielonka, Bolkart, Beeler, and Thies]{Zielonka2024GEM}
Wojciech Zielonka, Timo Bolkart, Thabo Beeler, and Justus Thies.
\newblock Gaussian eigen models for human heads.
\newblock \emph{arXiv:2407.04545}, 2024.

\bibitem[Zielonka et~al.(2025)Zielonka, Garbin, Lattas, Kopanas, Gotardo, Beeler, Thies, and Bolkart]{zielonka2025synshot}
Wojciech Zielonka, Stephan~J. Garbin, Alexandros Lattas, George Kopanas, Paulo Gotardo, Thabo Beeler, Justus Thies, and Timo Bolkart.
\newblock Synthetic prior for few-shot drivable head avatar inversion.
\newblock \emph{arXiv:2501.06903}, 2025.

\bibitem[Zollh{\"o}fer et~al.(2018)Zollh{\"o}fer, Thies, Garrido, Bradley, Beeler, P{\'e}rez, Stamminger, Nie{\ss}ner, and Theobalt]{Zollhfer2018StateOT}
Michael Zollh{\"o}fer, Justus Thies, Pablo Garrido, Derek Bradley, Thabo Beeler, Patrick P{\'e}rez, Marc Stamminger, Matthias Nie{\ss}ner, and Christian Theobalt.
\newblock State of the art on monocular 3d face reconstruction, tracking, and applications.
\newblock \emph{Computer Graphics Forum}, 37, 2018.

\bibitem[Zwicker et~al.(2001)Zwicker, Pfister, van Baar, and Gross]{Zwicker2001SurfaceS}
Matthias Zwicker, Hans~R{\"u}diger Pfister, Jeroen van Baar, and Markus~H. Gross.
\newblock Surface splatting.
\newblock \emph{Proceedings of the 28th annual conference on Computer graphics and interactive techniques}, 2001.

\end{thebibliography}
